\documentclass[10pt,twocolumn,letterpaper]{article}

\usepackage[pagenumbers]{cvpr} 

\usepackage[dvipsnames]{xcolor}
\usepackage{mwe} 
\usepackage{tabularx}
\usepackage{amsmath}
\usepackage{multirow}
\usepackage{siunitx}
\usepackage{makecell}
\usepackage{adjustbox}
\usepackage{cuted}

\renewcommand{\paragraph}[1]{\vspace{4pt}\noindent\textbf{#1}}

\definecolor{cvprblue}{rgb}{0.21,0.49,0.74}
\usepackage[pagebackref,breaklinks,colorlinks,citecolor=cvprblue]{hyperref}

\def\paperID{9803} 
\def\confName{CVPR}
\def\confYear{2024}

\title{Text2QR: Harmonizing Aesthetic Customization and Scanning Robustness for Text-Guided QR Code Generation}

\author{
Guangyang Wu\quad
\stepcounter{footnote}Xiaohong Liu\thanks{~Corresponding author.}\quad 
Jun Jia\quad
Xuehao Cui\quad
Guangtao Zhai\quad\\
Shanghai Jiao Tong University}

\begin{document}
\maketitle
\begin{strip}
\begin{minipage}{\textwidth}\centering
\vspace{-30pt}
\includegraphics[width=\linewidth]{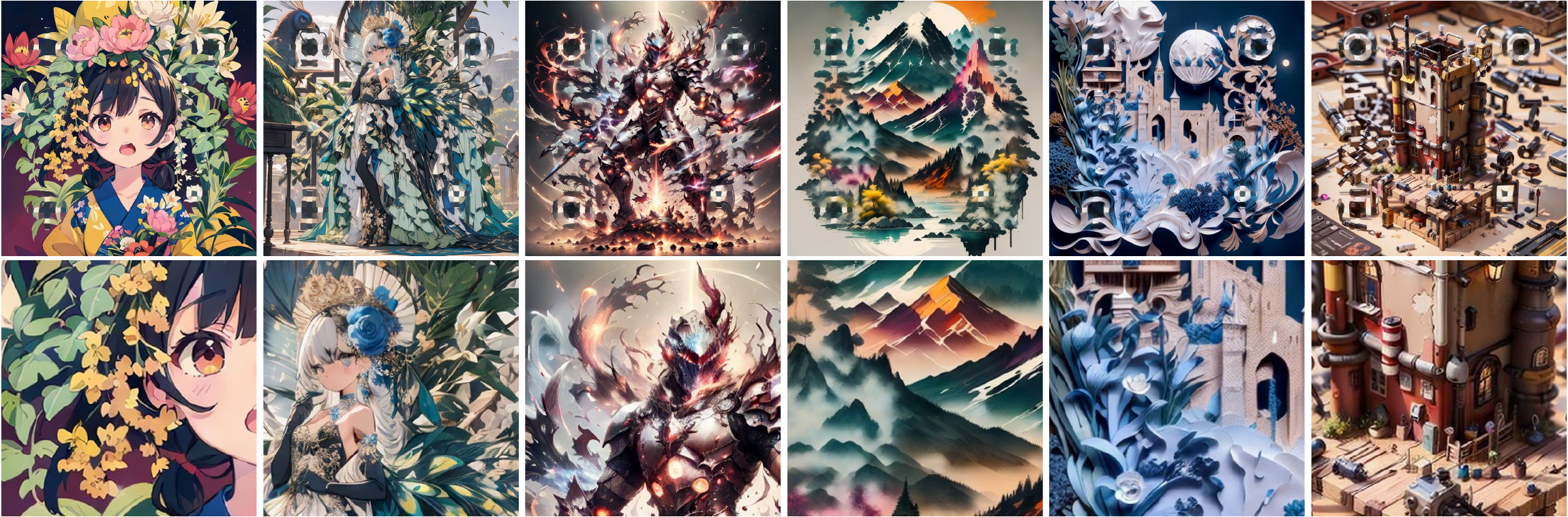}
\captionof{figure}{Aesthetic QR codes (first row) and their zoom-in counterparts (second row) generated by Text2QR. Our QR codes not only exhibit exceptional scanning robustness but also showcase allure and intricate details, accommodating a diverse range of customized styles.}
\label{figurelabel}
\end{minipage}
\end{strip}

\begin{abstract}\label{sec:abstract}
\vspace{-10pt}
    In the digital era, QR codes serve as a linchpin connecting virtual and physical realms. Their pervasive integration across various applications highlights the demand for aesthetically pleasing codes without compromised scannability. However, prevailing methods grapple with the intrinsic challenge of balancing customization and scannability. Notably, stable-diffusion models have ushered in an epoch of high-quality, customizable content generation. This paper introduces Text2QR, a pioneering approach leveraging these advancements to address a fundamental challenge: concurrently achieving user-defined aesthetics and scanning robustness.
    To ensure stable generation of aesthetic QR codes, we introduce the QR Aesthetic Blueprint (QAB) module, generating a blueprint image exerting control over the entire generation process. Subsequently, the Scannability Enhancing Latent Refinement (SELR) process refines the output iteratively in the latent space, enhancing scanning robustness. This approach harnesses the potent generation capabilities of stable-diffusion models, navigating the trade-off between image aesthetics and QR code scannability.
    Our experiments demonstrate the seamless fusion of visual appeal with the practical utility of aesthetic QR codes, markedly outperforming prior methods. Codes are available at \url{https://github.com/mulns/Text2QR}

\vspace{-10pt}
\end{abstract}
    
\section{Introduction}\label{sec:intro}

In an age where digital interaction seamlessly converges with the physical world, Quick Response (QR) codes serve as vital conduits connecting these realms~\cite{garateguy2014qr,stylized,picode,racode,su2021artcoder}. These ubiquitous two-dimensional codes have found extensive utility, bridging the divide between the physical and digital domains, yet their appearance remains stark, consisting of black and white modules engineered primarily for functional efficiency rather than aesthetic allure.

Amidst a growing consensus among users and stakeholders, a desire has arisen for QR codes that not only fulfill their core functions but also captivate with their visual appeal. The simplicity of QR codes, while undeniably efficient, is increasingly viewed as a missed opportunity to seamlessly integrate them into the modern visual landscape. The demand for aesthetically pleasing QR codes has proliferated, transcending boundaries into marketing, advertising, and artistic domains~\cite{stylized,picode,racode,su2021artcoder}.

Early techniques centered on image-to-image transformations, utilizing reshuffling~\cite{su2021artcoder}, fusion~\cite{artup,garateguy2014qr}, and style transfer methods~\cite{su2021artcoder,stylized}. Although effective in generating predefined image styles, these approaches struggled to accommodate the diverse stylistic preferences of users, leaving a gap for a unified solution that addresses both customization and consistency. Recent advancements in the intersection of image generation and control have marked a transformative era. stable-diffusion models~\cite{latent_diffusion,controlnet} have emerged as robust engines for producing high-quality, versatile, and dynamically ranged content. Concurrently, an innovative approach for aesthetic QR code generation surfaced~\cite{antfu}, leveraging ControlNet's capability to modulate luminance and darkness relationships within QR codes. However, this approach encountered a critical challenge, often exhibiting instability, necessitating the incorporation of auxiliary control models and manual parameter adjustments~\cite{antfu} to ensure both scannability and content quality.

In addressing these challenges, we introduce the innovative Text2QR pipeline, providing a solution for seamlessly generating QR codes that balance user-defined aesthetics and robust scannability. Our framework unfolds through three key steps: \textbf{(1)} Users initiate the process by generating their preferred images using the stable-diffusion model, while simultaneously encoding their desired message into a QR Code. \textbf{(2)} The synergy begins with the blending of these images in the QR Aesthetic Blueprint (QAB) module. This module generates a blueprint image, incorporating content from the pre-generated image (guidance image) and accurately reflecting the encoded message within the QR code. The blueprint image is then fed into ControlNet, guiding the stable-diffusion models to preserve user-defined aesthetics and maintain desired relationships among light and dark blocks of the QR code. While the generated results may pose decoding challenges, they exhibit a substantially improved distribution of light and dark blocks while remaining consistent with user preferences. \textbf{(3)} Subsequent to this stage, we construct an energy equation to quantify content and message consistency in the generated results. Optimizing this energy equation through gradient ascent iterations on latent codes gradually enhances scan robustness while preserving content consistency. Finally, the output QR code excels in both aesthetic appeal and scannability, achieving the delicate balance between user-defined customization and robust utility.

The contributions of this work can be summarized as:

\noindent$\bullet$ An integrated pipeline, Text2QR, that harmonizes user-defined aesthetics and robust scannability in QR code generation.

\noindent$\bullet$ The introduction of the QR Aesthetic Blueprint (QAB) for creating template images and the Scannability-Enhancing Latent Refinement (SELR) process for optimizing scan robustness while maintaining aesthetics.

\noindent$\bullet$ Superior performance compared to existing techniques, establishing Text2QR as a state-of-the-art solution for QR code generation that excels in both visual quality and scanning robustness.
\section{Related Works}\label{sec:related}


\paragraph{Aesthetic 2D Barcode.}
In the era of digital interaction, QR codes play a pivotal role in bridging the virtual and physical realms. A variety of aesthetically pleasing 2D barcodes have been proposed as alternatives to the less appealing QR codes. Halftone QR codes, proposed by Chu et al.~\cite{chu2013halftone}, rearrange the black/white modules of QR codes into an outline that semantically matches an input image. QR Image~\cite{garateguy2014qr, artup} leverages the redundancy in the coding rules of QR codes to embed color images within QR codes. Recently, Su et al.~\cite{su2021q,su2021artcoder} have combined QR codes with style transfer to create artistic QR codes. These methods adhere to the standard QR code encoding rules and can be scanned and decoded by a common mobile phone scanner.
To minimize the visibility of the locating patterns, Chen et al.~\cite{chen2018robust,racode,ma2023oacode} have designed encoding rules to satisfy the sensitivity of human visual system, making the locating patterns less noticeable. Additionally, TPVM~\cite{TPVM} hides QR codes in videos, utilizing the frame rate difference between screens and human eyes. Similarly, invisible information hiding is applied to make information invisible but decodable after camera shooting~\cite{fang2018screen,TERA,stegastamp,RIHOOP,wengrowski2019light,jia2022learning}.

\paragraph{Diffusion Based Generative Models.}
Deep learning-based image processing~\cite{RPSRMD,pred,fastllve,accflow,vsr,raw-vsr,griddehaze,griddehaze+,stsr,vfiformer,vfigdc} and generation methods~\cite{a3gan,ciagan,dalle,glide} has been fastly developed recently. Diffusion models such as GLIDE~\cite{glide}, DALLE-2~\cite{dalle}, Latent Diffusion \cite{latent_diffusion}, and Stable Diffusion \cite{latent_diffusion}, are proposed as a novel kind of generative model. These models create images through iterative denoising of initial random Gaussian noise and are able to outperform existing methods in many generative tasks. Of these, Stable Diffusion~\cite{latent_diffusion} is particularly innovative, as it transitions the denoising process from the image domain to a variational autoencoder's latent space, leading to significant reductions in data dimensions and training time. Alongside these advancements, various recent studies have presented methods for introducing diverse conditions to control the diffusion process. ControlNet~\cite{controlnet} and T2I-Adapter~\cite{t2i_adapter} focus on structural control, with the former introducing an adapter mirroring stable diffusion's structure and trained under structural conditions, while the latter fine-tunes a lightweight adapter for detailed control over the produced scenes and content from the diffusion model. Instead, BLIP-Diffusion~\cite{blip} and SeeCoder~\cite{SeeCoder} aim to achieve controllable results based on image style. 
BLIP-Diffusion~\cite{blip} extracts multi-modal topic representations and combines them with text prompts. While SeeCoder~\cite{SeeCoder} discards text prompts and use reference images as control parameters.

\section{Preliminary}
Prior to presenting our method, we elucidate the process by which a QR code scanner decodes binary information from an aesthetic QR code image. Given an colored image featuring a QR code, we initially transform it into a grayscale representation by extracting its luminance channel (Y-channel of the YCbCr color space), denoted as $I \in \mathbb{R}^{H\times W}$, encompassing $L$ gray levels (typically 256).
The scanner initially locates the Finder and Alignment markers~\cite{zxing,artup} to identify the QR code region and extract essential information such as the number of modules and module size. Let the QR code encompass $n\times n$ modules, each of size $a\times a$ pixels, where $n\cdot a \leq \min(H, W)$. Using the marker information, we construct a grid comprising $n^2$ modules denoted as $M_k, k\in[1,2,\dots,n^2]$. This grid divides image $I$ into $n^2$ patches, represented as $I_{M_k} \in \mathbb{R}^{a\times a}$.

The $k$-th module is decoded into a 1-bit information $\tilde{I}_k$, represented as 0 or 1, where $\tilde{I}\in \mathbb{R}^{n\times n}$ is the resulting binary image. Typically, scanners sample pixels within a central subregion of each module~\cite{artup,stylized}. Let $\theta$ be a square region with a size of $x\times x$ centered on module $M_k$, and $\mathbf{p}\in \{1,2,\dots,H\}\times \{1,2,\dots,W\}$ denotes pixel coordinates of $I$. The decoded binary value $\tilde{I}_k$ by a scanner is expressed as:
\begin{equation}\label{eq:decode}
    v_k = \frac{1}{x^2}\sum_{\mathbf{p}\in \theta} I_{M_k}(\mathbf{p}); \quad
    \tilde{I}_k =
    \begin{cases}
        0, \text{if } v_k \leq \mathcal{T}_b,\\
        1, \text{if } v_k \geq \mathcal{T}_w,\\
        -1, \text{otherwise}.
    \end{cases}
\end{equation}
Here, $\mathcal{T}_b$ and $\mathcal{T}_w$ are thresholds for binarization. To account for symmetry, we set $\mathcal{T}_b=L\cdot (1-\eta)/2$ and $\mathcal{T}_w=L\cdot (1+\eta)/2$, where $\eta\in(0,1)$. The hyperparameter $\eta$ governs the strictness of the binarization process.

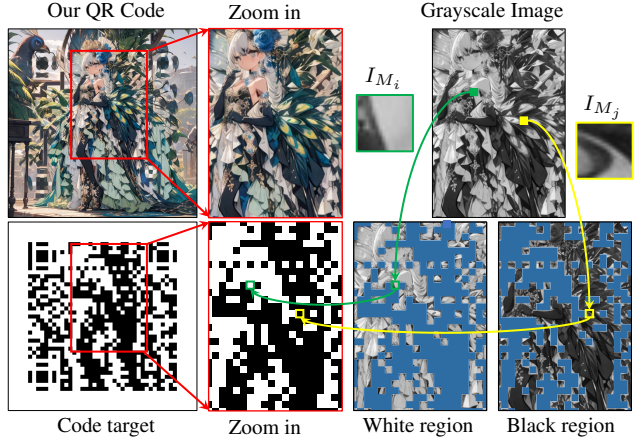
\begin{figure}
    \centering
    \includegraphics[width=\linewidth]{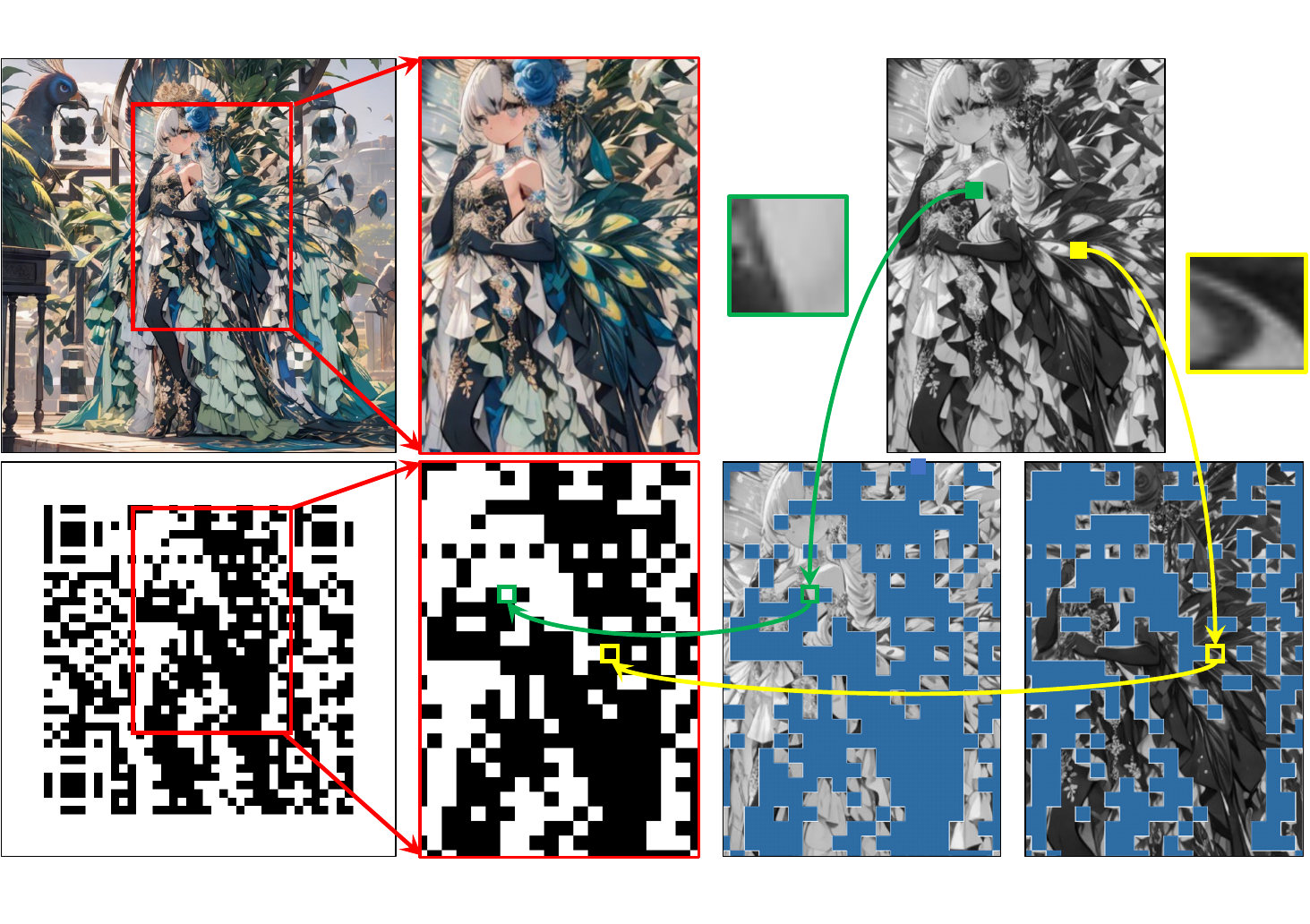}
    \begin{picture}(0,0)
        \put(-80,175){\makebox(0,0)[c]{\footnotesize Our QR Code}}
        \put(-80,18){\makebox(0,0)[c]{\footnotesize Code target}}
        \put(-20,175){\makebox(0,0)[c]{\footnotesize Zoom in}}
        \put(-20,18){\makebox(0,0)[c]{\footnotesize Zoom in}}
        \put(38,18){\makebox(0,0)[c]{\footnotesize White region}}
        \put(92,18){\makebox(0,0)[c]{\footnotesize Black region}}
        \put(66,175){\makebox(0,0)[c]{\footnotesize Grayscale Image}}
        \put(25,150){\makebox(0,0)[c]{\footnotesize $I_{M_i}$}}
        \put(108,139){\makebox(0,0)[c]{\footnotesize $I_{M_j}$}}
    \end{picture}
    \vspace{-15pt}
    \caption{Illustration of preserving scanning-robustness. Each module in our QR code (e.g. $I_{M_i}$ and $I_{M_j}$) is correspondingly mapped to white (green arrow) or black (yellow arrow) blocks, collectively forming a standard QR code target We use blue masks to filter the white and black modules for better visualization.}
    \label{fig:visualize}
    \vspace{-6pt}
\end{figure}

Conventional QR code markers, traditionally characterized by square patterns, have shown adaptability to diverse styles while maintaining readability for conventional QR code scanners, as indicated by recent studies~\cite{chen2018robust,ma2023oacode,antfu}. Achieving this adaptability involves specific pixel ratios, such as 1:1:3:1:1 for black and white modules, as detailed in~\cite{zxing,antfu}. This flexibility includes preserving a cross center region to convey relevant information. Moreover, within the data regions of QR codes, maintaining binary results despite variations in sampled pixel colors is crucial for scanning robustness. As shown in Figure~\ref{fig:visualize}, even when colors and shapes subtly blend and vary, the sampled pixels consistently yield binary results aligned with ideal QR codes, ensuring robust decoding by standard QR code readers.

To facilitate analysis, we define the probability $e(I)=p(\tilde{I}_k = \mathcal{M}_k)$ to assess the error level of image $I$ with respect to the code target $\mathcal{M}\in\mathbb{R}^{n\times n}$. Notably, the function $e$ only characterizes the error proportion within data regions, omitting Finder and Alignment regions in a QR code.

\section{Method}\label{sec:method}

\begin{figure*}[ht]
    \centering
    \includegraphics[width=\linewidth]{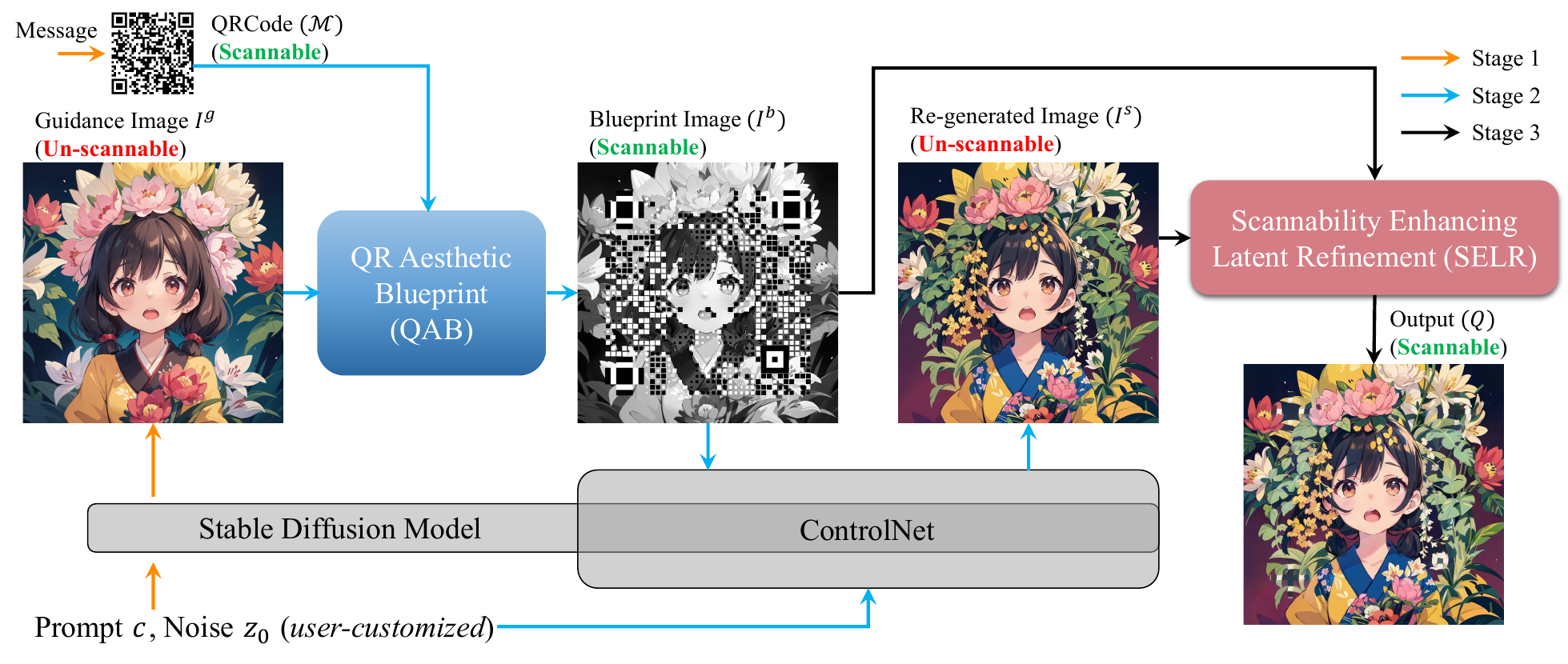}
    \caption{Overall Structure of the Text2QR. The pipeline consists of three stages, denoted with orange, blue and black lines. We propose the QAB module for generating a blueprint image used as controlling guidance, and propose the SELR module for refining the controlled output to enhance its scanning robustness.}
    \label{fig:overall}
\end{figure*}
\vspace{-2mm}

\subsection{Overall}

Figure~\ref{fig:overall} illustrates the overall structure of Text2QR, which is grounded in the Stable Diffusion (SD) model denoted as $\mathcal{G}$. The powerful customization capability enables the SD model to generate a user-preferred image $I^g = \mathcal{G}(c, z_0)$ with customized prompts $c$ and input noise $z_0$. Simultaneously, the input message is encoded into a QR code target $\mathcal{M}$, comprising $n\times n$ binary values (1 for white, 0 for black), representing the ideal color of each module.

Given $I^g$ and $\mathcal{M}$, Text2QR is designed to yield an aesthetically pleasing QR code denoted as $Q$. In pursuit of visual allure, $Q$ faithfully mirrors the semantic content, aesthetic style, and figure layout inherent in $I^g$. Simultaneously, for practical functionality, $Q$ is engineered to seamlessly reveal the encoded message upon scanning, adaptive to any standard QR code reader. The architectural framework of our pipeline unfolds across three distinct stages:

In the first stage, users prepare $I^g$ and $\mathcal{M}$, recording the associated parameters ($c$ and $z_0$). During the second stage, a pivotal step entails the seamless integration of information encapsulated within $I^g$ and $\mathcal{M}$ to formulate a comprehensive blueprint image, denoted as $I^b$, through the innovative QR Aesthetic Blueprint (QAB) module. Subsequently, $I^b$ undergoes processing in a ControlNet $\mathcal{C}$ to exert influence on the SD model. This influence involves adjusting intermediate features through a controlled process defined as:
\begin{equation}
    I^s = \mathcal{G}(c, z_0 | \mathcal{C}(I^b, c, z_0)).
\end{equation}
This integration ensures a synergistic output $I^s$ that harmoniously balances the aesthetic preferences derived from $I^g$ with the structural constraints imposed by $\mathcal{M}$. In the conclusive stage, $I^s$ undergoes iterative fine-tuning through the Scannability Enhancing Latent Refinement (SELR) module. This refines the scanning robustness of $I^s$ while meticulously preserving its aesthetic qualities. The output of this process is an aesthetically impressive QR code $Q$. Subsequent sections delve into detailed expositions on both QAB and SELR modules.

\subsection{QR Aesthetic Blueprint}
The module aims to create a scannable blueprint by integrating QR code information and guidance image details. Initially, we extract the luminance channel, denoted as $I^g_y$, from the guidance image $I^g$. To ensure comparable distributions, we preprocess $I^g_y$ and $\mathcal{M}$ using histogram polarization for luminance adjustment and a module reorganization method for pixel rearrangement, respectively.  Finally, the Adaptive-Halftone method is applied to blend them, yielding the blueprint image $I^b$.

\paragraph{Histogram polarization.} 
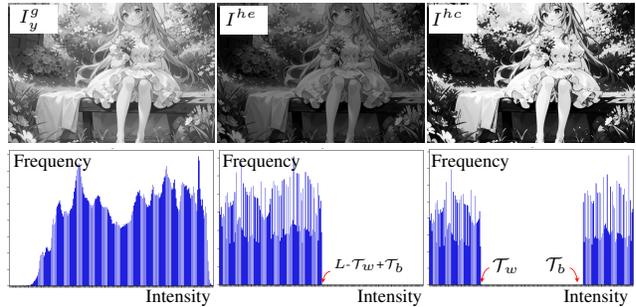
\begin{figure}
    \centering
    \includegraphics[width=\linewidth]{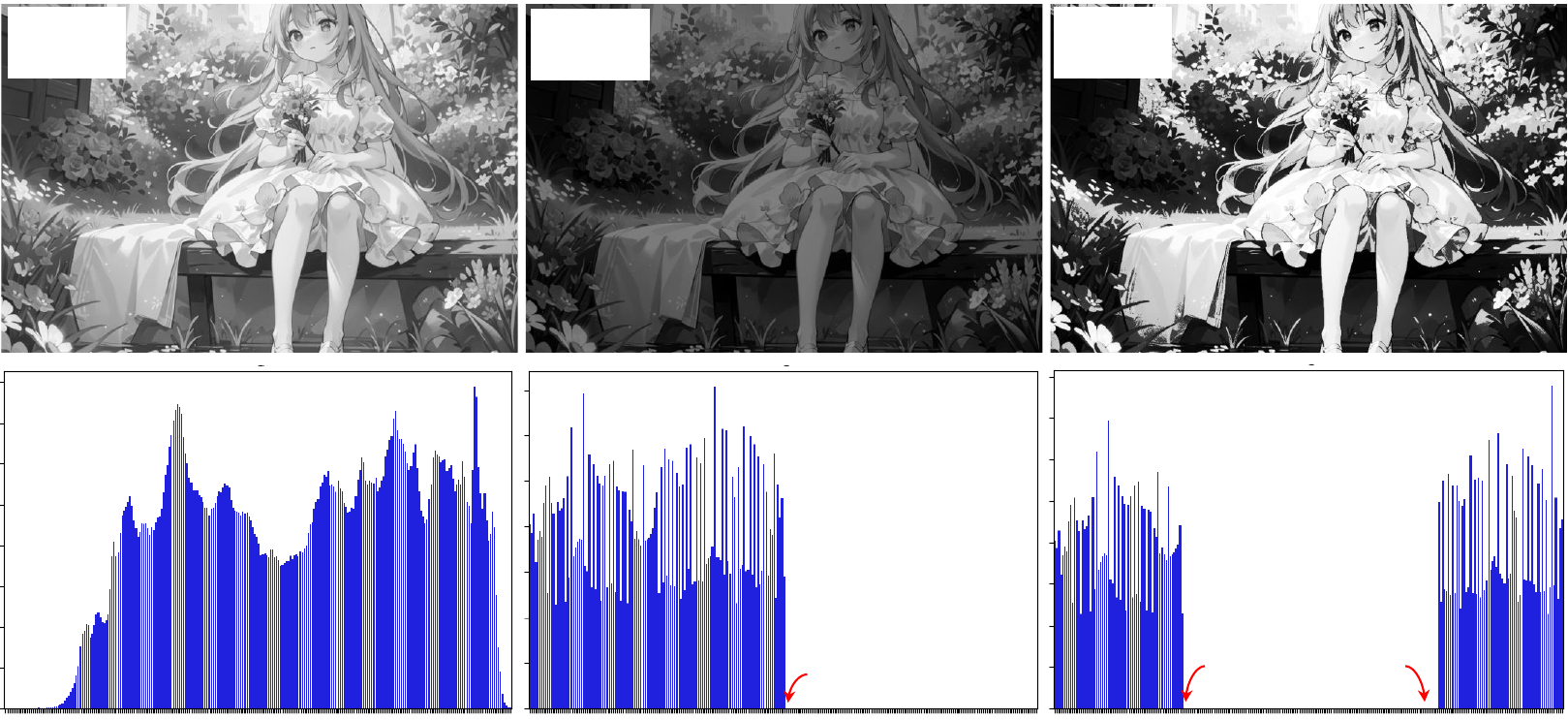}
    \begin{picture}(0,0)
        \put(-113,114){\makebox(0,0)[l]{\scriptsize $I_y^g$}}
        \put(-115,60){\makebox(0,0)[l]{\scriptsize Frequency}}
        \put(-35,115){\makebox(0,0)[l]{\scriptsize $I^{he}$}}
        \put(-36,60){\makebox(0,0)[l]{\scriptsize Frequency}}
        \put(43,115){\makebox(0,0)[l]{\scriptsize $I^{hc}$}}
        \put(43,60){\makebox(0,0)[l]{\scriptsize Frequency}}
        \put(93,9){\makebox(0,0)[l]{\scriptsize Intensity}}
        \put(15,9){\makebox(0,0)[l]{\scriptsize Intensity}}
        \put(-65,9){\makebox(0,0)[l]{\scriptsize Intensity}}
        \put(6,21){\makebox(0,0)[l]{\tiny $L$-$\mathcal{T}_w$+$\mathcal{T}_b$}}
        \put(66,21){\makebox(0,0)[l]{\scriptsize $\mathcal{T}_w$}}
        \put(86,21){\makebox(0,0)[l]{\scriptsize $\mathcal{T}_b$}}
    \end{picture}
    \vspace{-10pt}
    \caption{Visualization of the process of Histogram Polarization.}\label{fig:hist}\vspace{-10pt}
\end{figure}
The primary aim of this module is to harmonize the histogram distribution of $I^g_y$ with that of the QR code. This process enhances the contrast of $I^g_y$, yielding a high-contrast grayscale image $I^{hc}$.
The histogram polarization operation is represented by a look-up table $\mathcal{H}$, which maps pixel values from one gray level to another. For each pixel $\mathbf{p}$, let $\tau = I^g_y(\mathbf{p})$ and $\tau' = I^{hc}(\mathbf{p})$, we express this transformation as follows:
\begin{equation}
    \tau' = \mathcal{H}(\tau).
\end{equation}
Let $n_\tau$ denote the occurrences of gray level $\tau\in [0,L)$, we introduce the Cumulative Distribution Function (CDF) corresponding to gray level $\tau$ as:
\begin{equation}
    \text{cdf}(\tau) = \sum_{i=0}^{\tau}\frac{n_i}{H\times W}.
\end{equation}
The objective is to generate $I^{hc}$ with a flat histogram in the data range $[0, \mathcal{T}_b) \bigcup [\mathcal{T}_w, L)$, while excluding occurrences in the data range $[\mathcal{T}_b, \mathcal{T}_w)$. To achieve this, we first create a new image $I^{he}$, with a linearized CDF across the value range $[0,L - \mathcal{T}_w + \mathcal{T}_b]$. Let $\tilde{\tau} = I^{he}(\mathbf{p})$, we have:
\begin{equation}
    \tilde{\tau} = (L-\mathcal{T}_w+\mathcal{T}_b)\cdot \text{cdf}(\tau).
\end{equation}
Subsequently, we shift the pixels within the value range $[\mathcal{T}_b, L)$ by adding $\mathcal{T}_w - \mathcal{T}_b$ to obtain $I^{hc}$:
\begin{equation}
    \tau' = \mathcal{H}(\tau) = 
    \begin{cases}
        \tilde{\tau}, & \text{if } \tilde{\tau} < \mathcal{T}_b, \\
        \tilde{\tau} + \mathcal{T}_w-\mathcal{T}_b, & \text{if } \tilde{\tau} \geq \mathcal{T}_b.
    \end{cases}
\end{equation}
Figure~\ref{fig:hist} shows visualization of these processes, illustrating the transformation of the image with a polarized histogram and high-contrast luminance.

\paragraph{Module reorganization.}
To blend the QR code $\mathcal{M}$ with $I^{hc}$, we first binarize the $I^{hc}$ to a binary image $I^{bin}$. This binary image guides the module reorganization method, denoted as $\mathcal{E}_r$, which rearranges the modules of $\mathcal{M}$ while keeping the encoded information. The process can be formulated as:
\begin{gather}
    I^{bin}({\mathbf{p}}) = 
    \begin{cases}
        0, & \text{if } I^{hc}({\mathbf{p}}) < \mathcal{T}_b, \\
        1, & \text{if } I^{hc}({\mathbf{p}}) > \mathcal{T}_w,
    \end{cases}\\
    \mathcal{M}^r = \mathcal{E}_{r}(\mathcal{M}, I^{bin}).
\end{gather}

\paragraph{Adaptive-Halftone blending.}
\begin{figure}
    \centering
    \includegraphics[width=\linewidth]{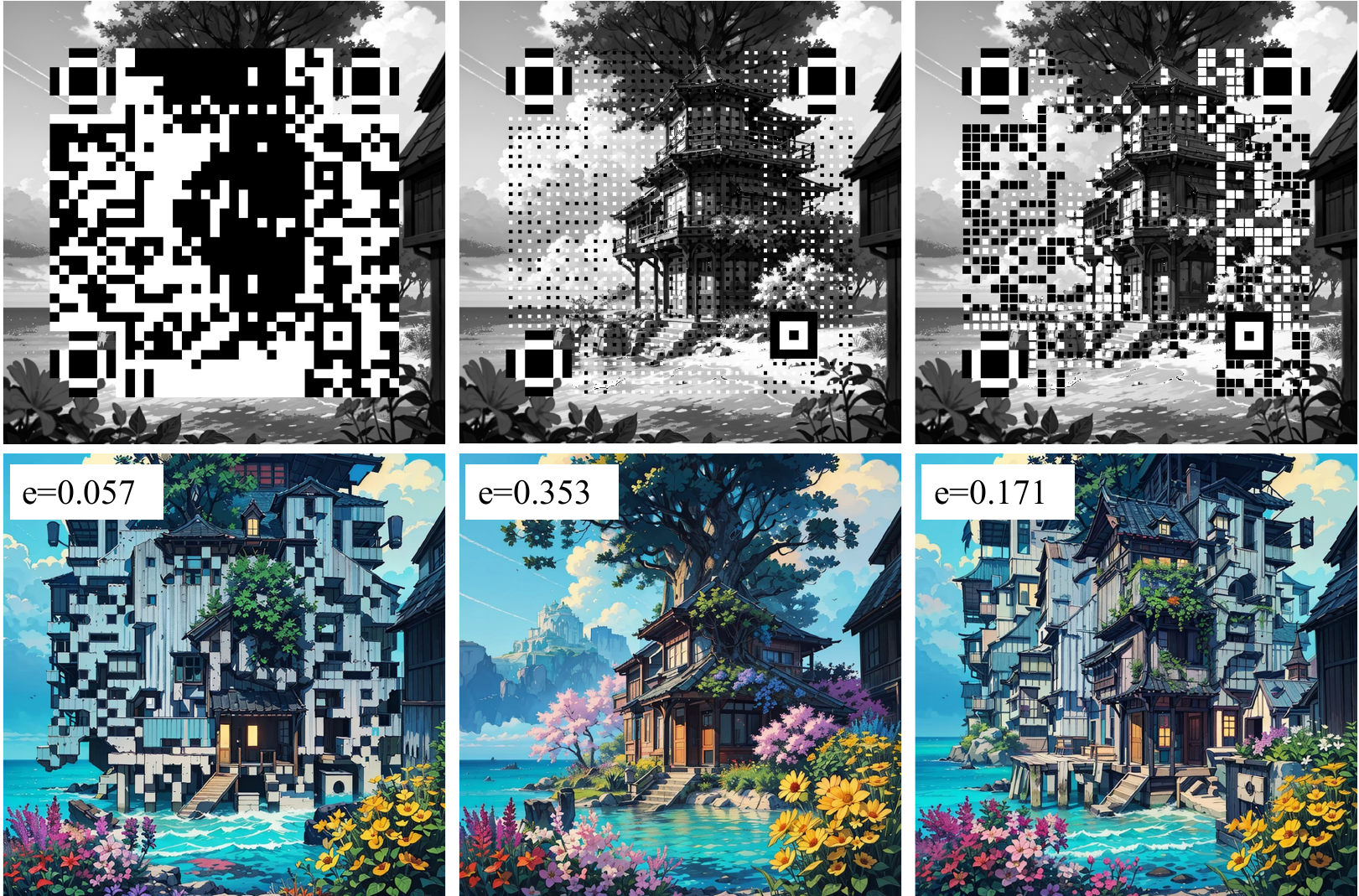}
    \caption{Comparison of blueprint images and their corresponding ControlNet output. Utilizing a pure QR code as the blueprint (first column) yields a low error level $e$ but lacks semantic features. Employing a fixed size of $u=\frac{a}{3}$ (second column) leads to a substantial error level. Our Adaptive-Halftone blending method preserves realistic image content with a minimal error level.}\label{fig:halftone}
    \vspace{-10pt}
\end{figure}
Considering the $k$-th module region $M_k$, we input an image patch after histogram polarization $I^{hc}_{M_k}$ and a target value $\mathcal{M}^r_{k} = 0$ or $1$. Our goal is to obtain the blueprint image $I^b$, where $I^b_{M_k}$ can be \textit{decoded to the correct information while preserving as much image content as possible}.

To achieve this objective, we introduce a novel blending method, Adaptive-Halftone blending. Specifically, for each module $M_k$, let $\theta_k$ be a square region of size $u\times u$ centered on the image patch $I_{M_k}^{hc}$ ($u\leq a$). We fill the $I^{hc}_{\theta_k}$ with value $\mathcal{M}^r_{k}$ to generate $I^b_{M_k}$. The square region size $u$ is optimized by minimizing the code distance within this module. The simulating decoded value of this module corresponding to $u$ is defined as $E_k(I^b_{M_k}|u)$:
\begin{align}
    E_k(I^b_{M_k}|u) &= \frac{1}{a^2} [ \sum_{\mathbf{p}\in \theta_k}L\cdot\mathcal{M}^r_{k} + \sum_{\mathbf{p}\notin \theta_k}I^b_{M_k}(\mathbf{p}) ] \\ \nonumber
    & = \frac{1}{a^2} [ u^2\cdot L\cdot\mathcal{M}^r_{k} + \sum_{\mathbf{p}\notin \theta_k}I^b_{M_k}(\mathbf{p}) ].
\end{align}
Particularly, $E_k(I^b_{M_k}|u) = L\cdot \mathcal{M}^r_{k}$ when $u=a$. The objective is to minimize the code distance:
\begin{equation}
    \label{eq:energy}
    u_k = \arg\min_u \|  E_k(I^b_{M_k}|u) - L\cdot (\eta\cdot \mathcal{M}^r_{k}+\frac{1-\eta}{2}) \|.
\end{equation}
According to the definition of thresholds $\mathcal{T}_b$ and $\mathcal{T}_w$, Equation~\ref{eq:energy} can be further simplified as:
\begin{equation}
    s_k = 
    \begin{cases}
        \arg\min_s \| E_k(I^b_{M_k}|u) - \mathcal{T}_b \|, \text{if } \mathcal{M}^r_k = 0, \\
        \arg\min_s \| E_k(I^b_{M_k}|u) - \mathcal{T}_w \|, \text{if } \mathcal{M}^r_k = 1.
    \end{cases}
\end{equation}

Having populated each module with a co-centered square block of adaptable size, we proceed to affix markers, including Finders and Alignment markers, onto the finalized blueprint image $I^b$. Figure~\ref{fig:halftone} showcases diverse $I^s$ generated from distinct blueprint images. Our method dynamically adjusts the size of the central block for each module in $I^b$. This adjustment involves shrinking the block size when $I^{hc}$ effectively encapsulates the module's information to preserve more image content. Conversely, it enlarges the block size when a more pronounced control signal is necessary to ensure the module's scannability in $I^s$.

\subsection{Scannability Enhancing Latent Refinement}
\begin{figure}
    \centering
    \includegraphics[width=\linewidth]{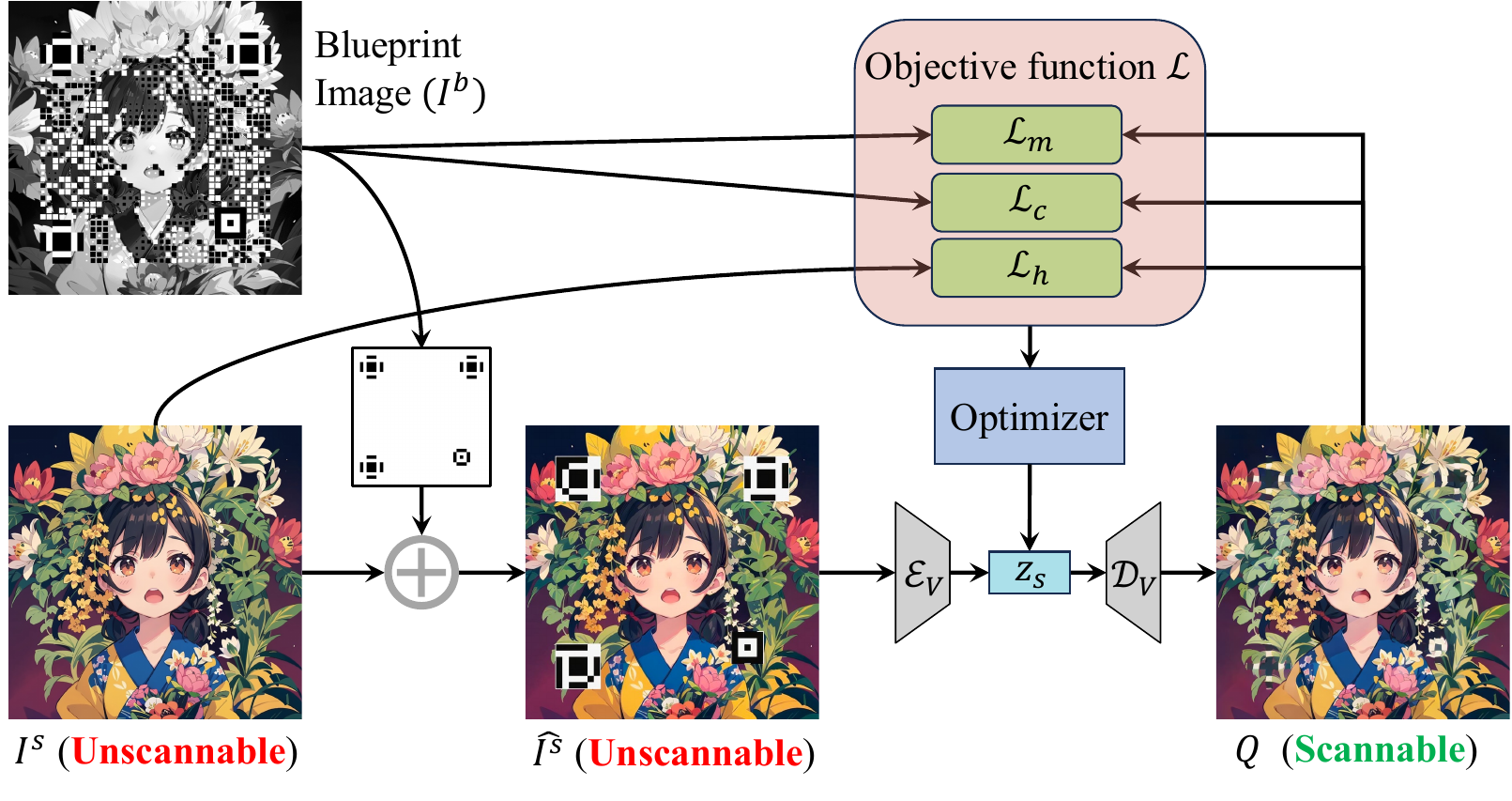}
    \caption{The recurrent pipeline of SELR.}\label{fig:sela}
    \vspace{-5pt}
\end{figure}
While the re-generated image $I^s$ adheres to the structural constraints imposed by $\mathcal{M}^r$, it often lacks scannability due to the presence of numerous error modules. Addressing this issue, the Scannability Enhancing Latent Refinement (SELR) module offers a meticulous refinement process to enhance scanning robustness. The markers, encompassing finder and alignment patterns, are pivotal for determining the location and angle of a QR code, thereby influencing its scannability. Hence, we integrate their appearances onto $I^s$ before refinement, denoted as $\widehat{I^s}$.
As shown in Figure~\ref{fig:sela}, we encode the augmented image $\widehat{I^s}$ into a latent code $z_s$ using the encoder of a pre-trained Variational AutoEncoder (VAE) model, denoted as $\mathcal{E}_{V}$. The total objective function $\mathcal{L}$ is defined as the weighted sum of three terms: marker loss $\mathcal{L}_{m}$, code loss $\mathcal{L}_{c}$, and harmonizing loss $\mathcal{L}_{h}$:
\begin{align}
\mathcal{L}(z) &= \lambda_1 \mathcal{L}_{m}(\mathcal{D}_{V}(z), I^b)        
                \nonumber\\
                &+ \lambda_2 \mathcal{L}_{c}(\mathcal{D}_{V}(z), I^b) \nonumber\\
                &+ \lambda_3 \mathcal{L}_{h}(\mathcal{D}_{V}(z), I^s),
\end{align}
where $\lambda_1$ to $\lambda_3$ are used to balance the multiple objectives. Here, $\mathcal{D}_{V}$ represents the decoder in the VAE model. The latent feature $z$ is initialized with $z_s$ and fine-tuned through an optimization process to minimize the total objective function, thereby controlling the scannability and aesthetic quality of the generated QR code $Q=\mathcal{D}_V(z)$. The code loss $\mathcal{L}_{c}$ is derived from the methodology proposed in~\cite{su2021artcoder}. This approach employs the SSLayer to extract module values and computes the module-based code loss through a competitive mechanism.

\paragraph{Marker loss.}
In line with previous discussions, scanners identify QR codes based on specific pixel ratios in marker regions. Consequently, our strategy centers on constraining the cross-center region of the marker, recognizing its crucial role in preserving scannability. To implement this, we introduce a binary mask, $\mathcal{K}_{cc}$, tailored to filter the cross-center region of markers. The aim is to safeguard the essential marker features against potential compromise due to aesthetic customization. Formally, the marker loss function $\mathcal{L}_{m}$ is defined as follows:
\begin{equation}
\mathcal{L}_{m}(Q, I^b) = \mathcal{K}_{cc} \cdot \parallel Q_y - I^b \parallel^2,
\end{equation}
where $Q_y$ denotes the luminance channel of the QR code $Q$. This formulation ensures marker integrity preservation while allowing for aesthetic modifications in non-marker regions of the QR code.

\paragraph{Harmonizing loss.}
Having addressed the marker and code-related scannability concerns, our approach further ensures the preservation of aesthetic qualities through a harmonizing loss.
To maintain the intrinsic aesthetic style of the generated QR code $Q$, we employ a harmonizing loss focused on optimizing visual quality while upholding its original appeal. This loss function computes the $L^2$-Wasserstein distance, denoted as $W_2$, between the feature maps of $Q$ and $I^s$. Specifically, feature map $f_i$ is extracted from $i$-th layer of a pre-trained VGG-19 network ($i \in [1, 6, 11, 18, 25]$). The loss is formulated as:
\begin{equation}
    \mathcal{L}_{h} = \sum_i W_{2}(f_i(Q), f_i(I^s)),
\end{equation}
where $f_i$ denotes the feature map extracted from the $i$-th layer ($i \in [1, 6, 11, 18, 25]$) of the VGG network. The $L^2$-Wasserstein distance, assuming the feature distributions approximate Gaussian distributions described by means and co-variances, can be expressed in a closed form~\cite{Wasserstein}. Let $P_1$ and $P_2$ be Gaussian measures on $\mathbb{R}^n$ with means $\mu_1$ and $\mu_2 \in \mathbb{R}^n$ and non-singular covariance matrices $C_1$ and $C_2 \in \mathbb{R}^{n\times n}$, respectively. The $L^2$-Wasserstein distance $W_2(P_1, P_2)$ is given by:
\begin{align}
    &A = trace(C_1+C_2-2(\sqrt{C_1} C_2 \sqrt{C_1})^\frac{1}{2}), \nonumber \\
    &W_2(P_1, P_2) = \sqrt{\parallel \mu_1-\mu_2 \parallel^2 + A}.
\end{align}
The integration of the harmonizing loss ensures that the optimized output not only meets the functional requirements but also preserves aesthetic qualities. In essence, the SELR module leverages marker, code, and harmonizing losses to optimize both the scannability and aesthetic appeal of the generated QR code.
\section{Experiments}\label{sec:exp}
We evaluate the performance of our QR codes in two aspects, aesthetic quality and scanning-robustness.

\renewcommand\theadfont{\bfseries}

\begin{table}[b]
    \caption{Average scanning success rates are assessed across various scanners, considering different sizes and angles. ``Scanner'' denotes the native scanner of each system. We compare the accuracy (\%) of our method and ArtCoder~\cite{su2021artcoder}.}
    \label{table:scannability}
    \centering
    \begin{tabularx}{\linewidth}{c c *{6}{>{\centering\arraybackslash}X}}
    \toprule
    \multirow{3}{*}{\thead{Mobile\\Phone}} & \multirow{3}{*}{\thead{APPs}} & \multicolumn{6}{c}{\thead{Accuracy (\%)}}  \\ 
    \cmidrule(lr){3-8}
    &   & \multicolumn{2}{c}{(3cm)$^2$} & \multicolumn{2}{c}{(5cm)$^2$} & \multicolumn{2}{c}{(7cm)$^2$}  \\ 
    
    \cmidrule(lr){3-4} \cmidrule(lr){5-6} \cmidrule(lr){7-8}
    &   & {$45^{\circ}$} & {$90^{\circ}$} & {$45^{\circ}$} & {$90^{\circ}$}   & {$45^{\circ}$} & {$90^{\circ}$}   \\ 
    
    \midrule
    \multirow{3}{*}{\renewcommand\theadfont{} \thead{iPhone\\14Pro}} & Scanner & 100 & 100 & 100 & 100 & 100 & 100 \\ 
    & TikTok &100 & 100 & 100 & 100 & 100 & 100 \\ 
    & WeChat &100 & 100 & 100 & 100 & \underline{96} & \underline{96}  \\ 
    \midrule
    \multirow{3}{*}{\renewcommand\theadfont{} \thead{Huawei\\P40}} & Scanner &100 & 100 & 100 & 100 & 100 & 100 \\ 
    & TikTok &100 & 100 & 100 & 100 & 100 & 100 \\ 
    & WeChat &100 & 100 & \underline{96} & 100 & \underline{96}& 100 \\
    \bottomrule
    \end{tabularx}
\end{table}

\subsection{Implementation}

We implement our program in PyTorch and conduct experiments on a NVIDIA GeForce 3090 GPU. For scanning-robustness assessment, we display QR codes on a 27-inch, 144Hz IPS-panel monitor. Default settings include $\eta=0.6$ (following the work~\cite{su2021artcoder}) and a controlling strength of 1.4 for the pre-trained ControlNet (i.e., QR-Monster~\cite{antfu}). The VAE model aligns with the SD model, featuring frozen parameters, and VGG-19 that is pre-trained on MS-COCO extracts feature maps in SELR. During refinement, we employ Adam optimizer for 400 iterations with the learning rate set to 0.002 and default weights $\lambda_1$, $\lambda_2$ and $\lambda_3$ set to 1.0. QR codes ($\mathcal{M}$) are generated in version 5 of size $592\times 592$ (i.e., $37\times 37$ modules, each of $16\times 16$~\cite{su2021artcoder}). In this paper, we conduct comparisons on dataset comprising 100 generated images of $1,024\times 1,024$, span various visual content and artistic styles.

\newlength{\myfigwidth}
\setlength{\myfigwidth}{0.16\textwidth} 
\begin{table*}[h]
    \caption{Visual comparison of different methods. More results can be found in the supplementary material.}
    \label{tab:comparison}
    \centering
    \begin{tabularx}{\linewidth}{>{\centering\arraybackslash}X *{6}{>{\centering\arraybackslash}X}}

    \toprule
    Input & QArt~\cite{qart} & {Halftone QR~\cite{chu2013halftone}} &  ArtCoder~\cite{su2021artcoder} & Quick QR~\cite{quickqr} & Text2QR (ours) \\
    \midrule

    \includegraphics[width=\myfigwidth, height=\myfigwidth]{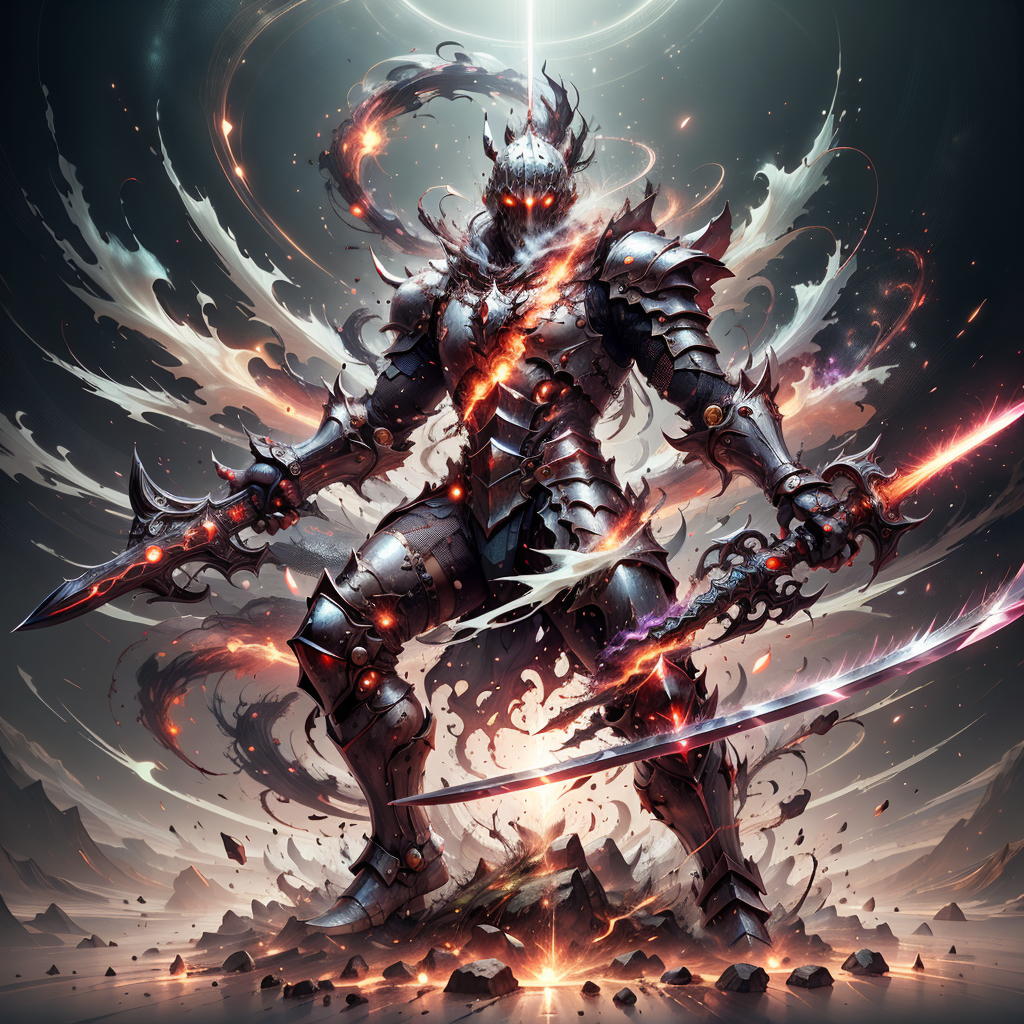} & 
    \includegraphics[width=\myfigwidth, height=\myfigwidth]{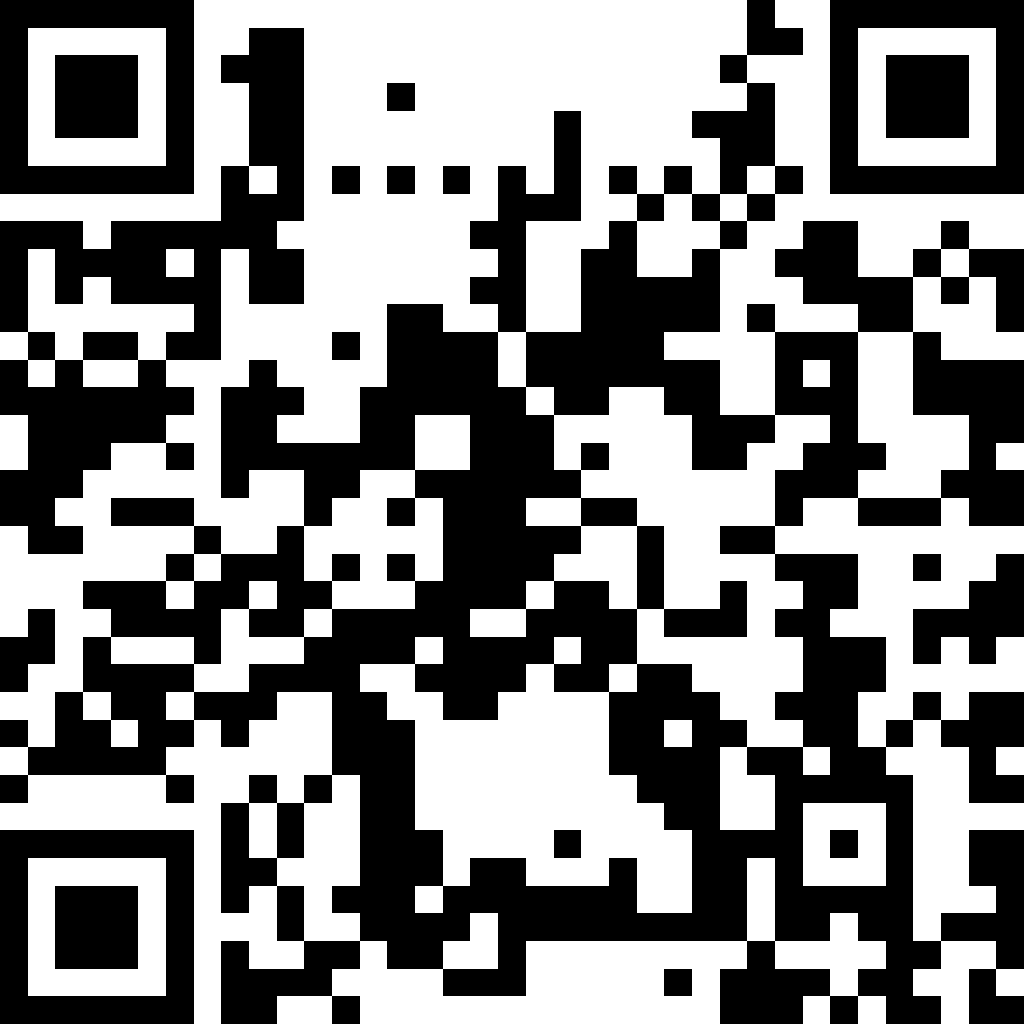} & 
    \includegraphics[width=\myfigwidth, height=\myfigwidth]{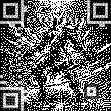} & 
    \includegraphics[width=\myfigwidth, height=\myfigwidth]{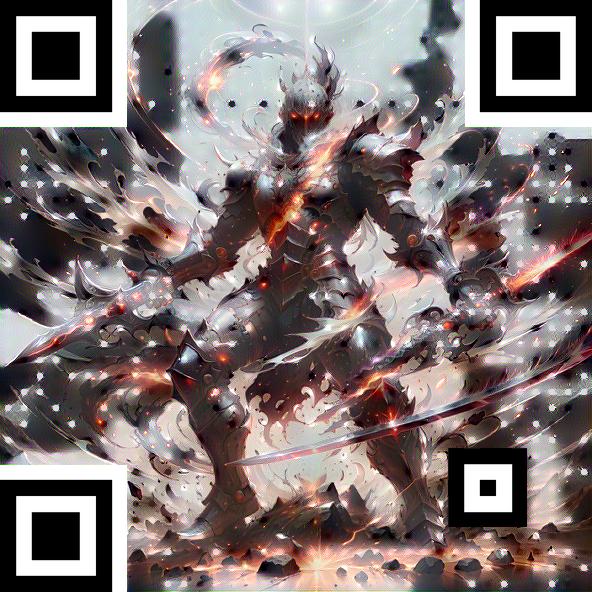} & 
    \includegraphics[width=\myfigwidth, height=\myfigwidth]{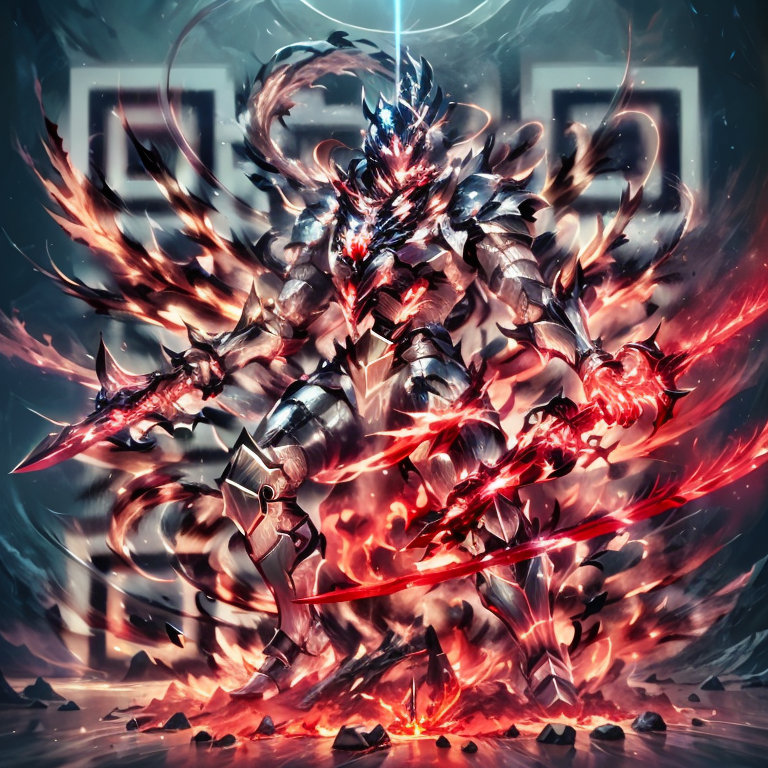} & 
    \includegraphics[width=\myfigwidth, height=\myfigwidth]{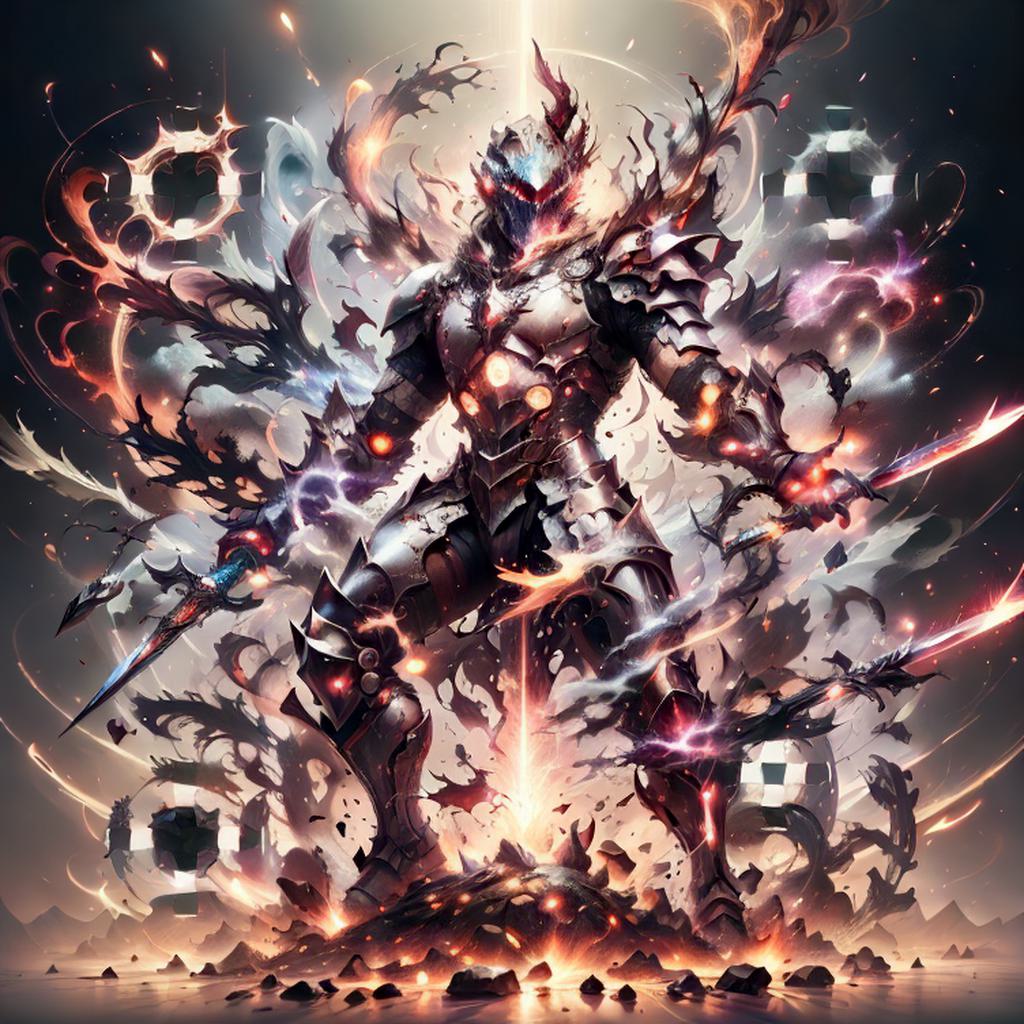} \\

    \includegraphics[width=\myfigwidth, height=\myfigwidth]{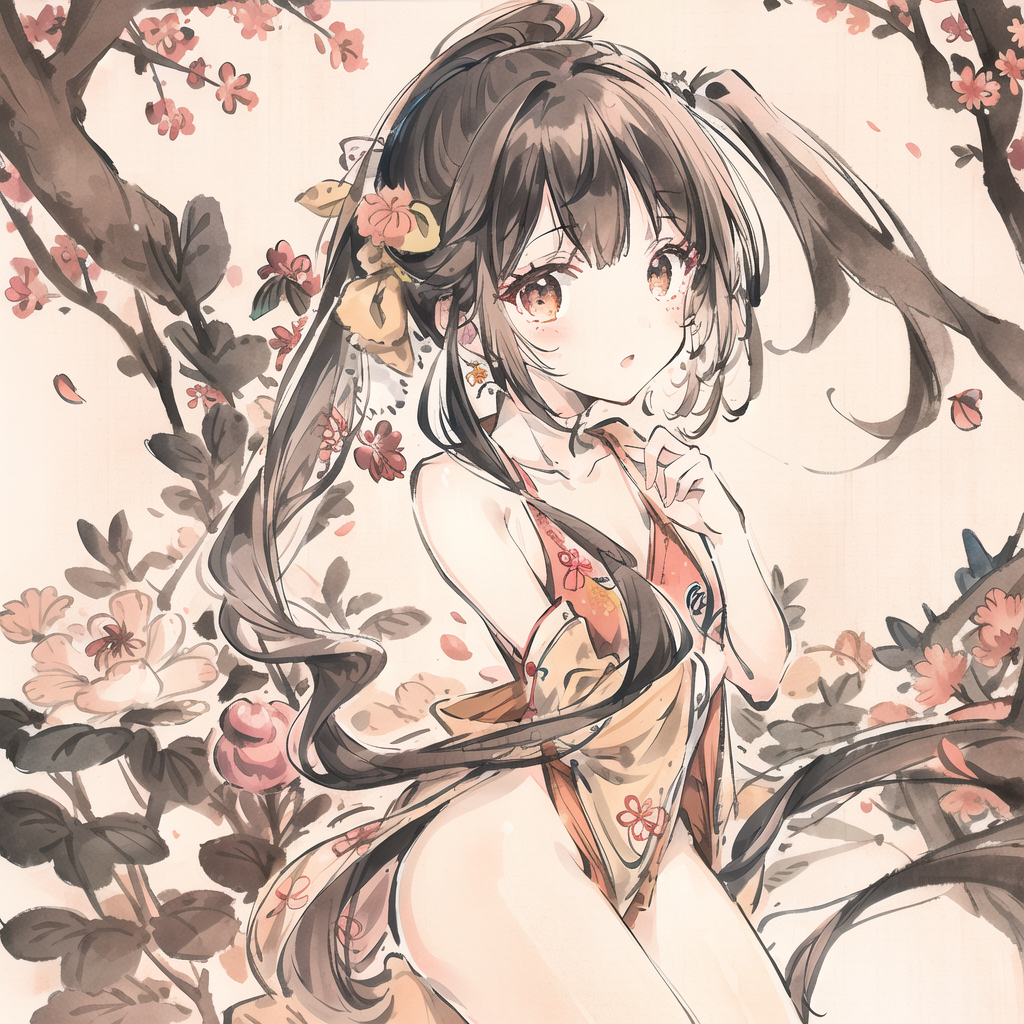} & 
    \includegraphics[width=\myfigwidth, height=\myfigwidth]{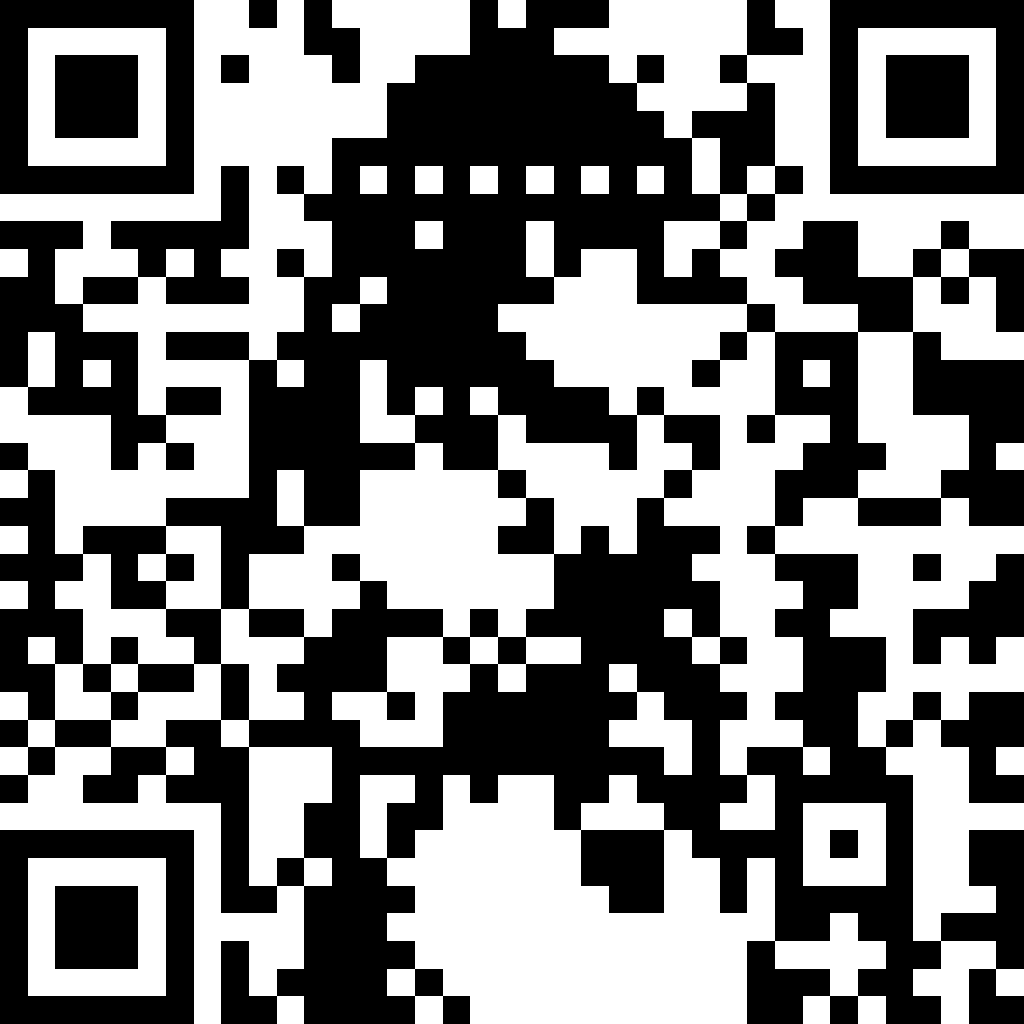} & 
    \includegraphics[width=\myfigwidth, height=\myfigwidth]{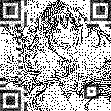} & 
    \includegraphics[width=\myfigwidth, height=\myfigwidth]{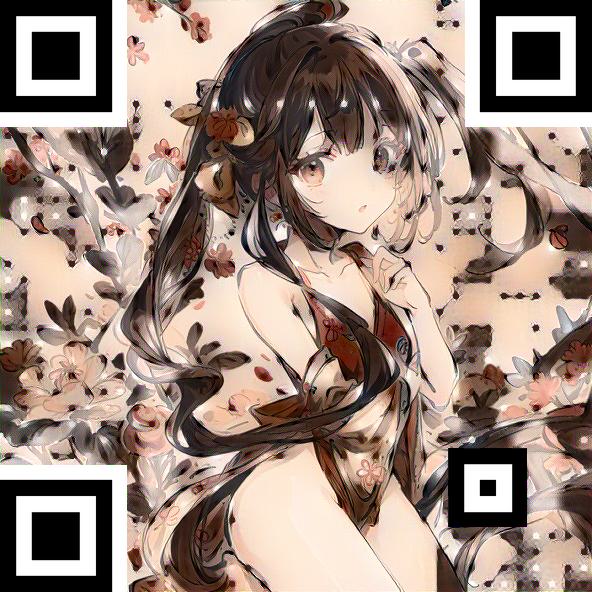} & 
    \includegraphics[width=\myfigwidth, height=\myfigwidth]{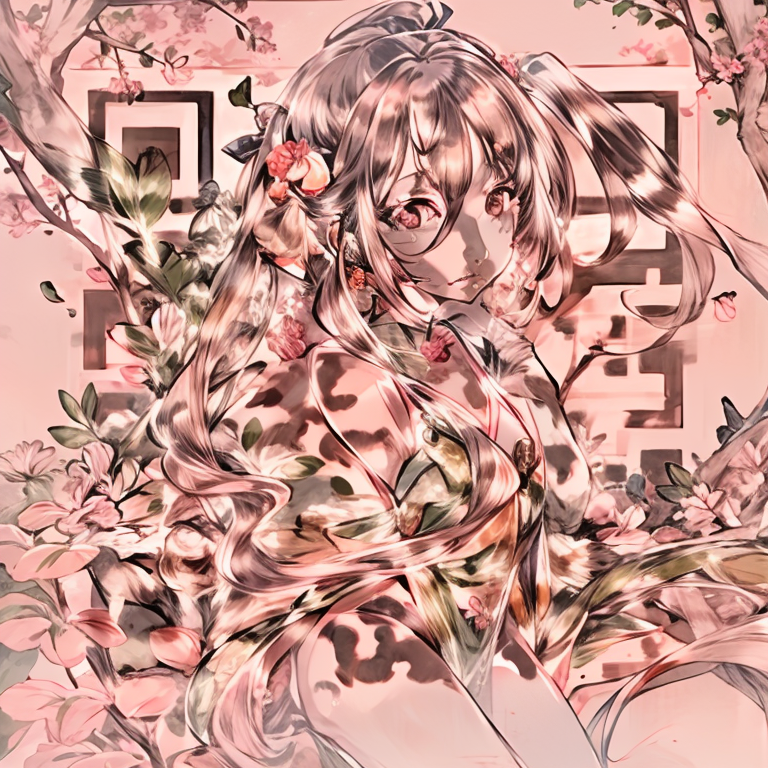} & 
    \includegraphics[width=\myfigwidth, height=\myfigwidth]{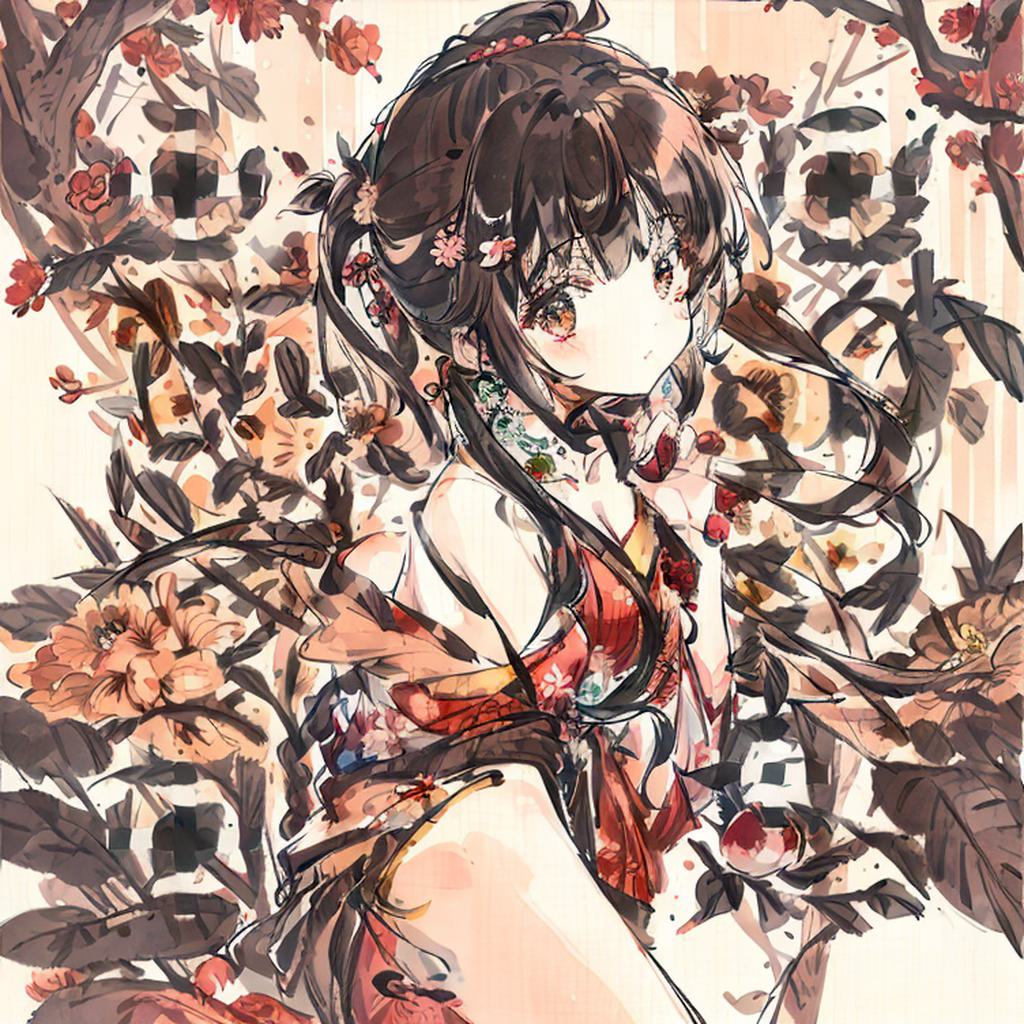} \\

    \includegraphics[width=\myfigwidth, height=\myfigwidth]{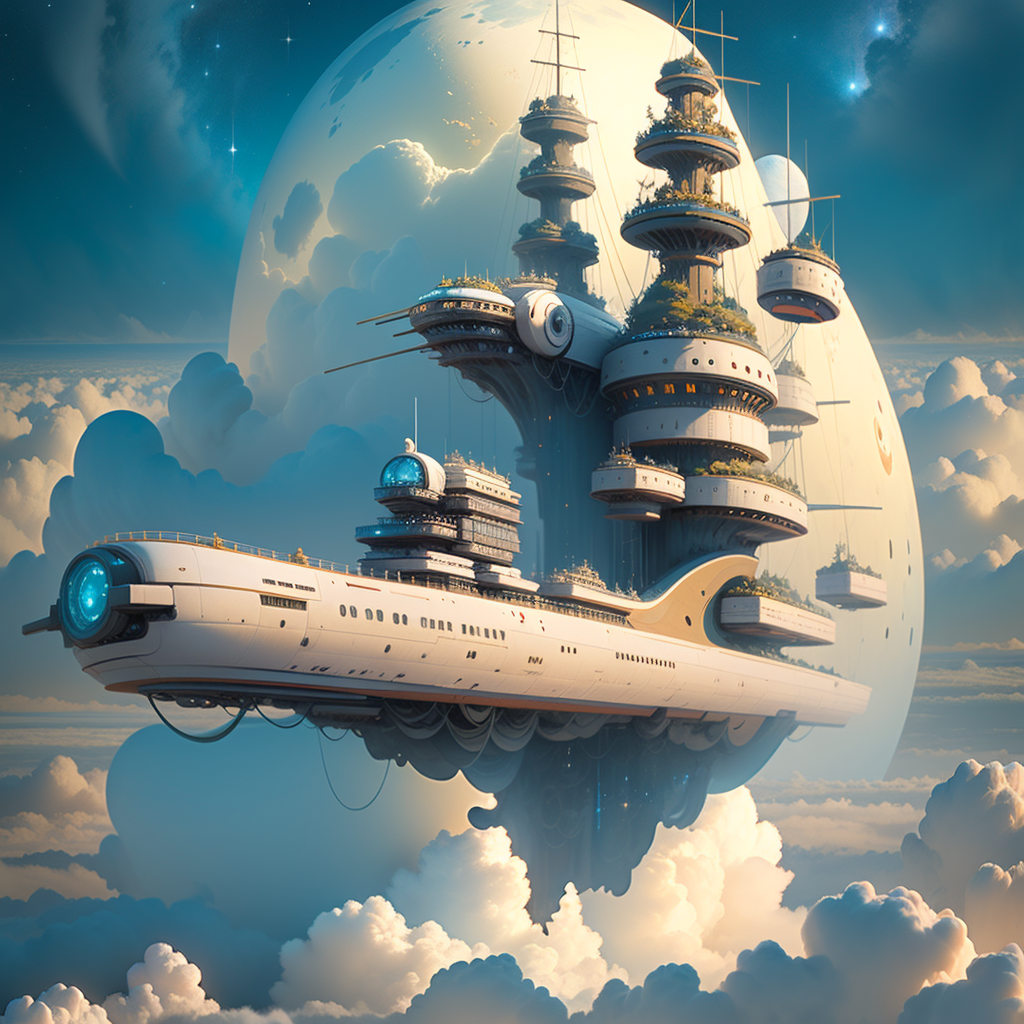} & 
    \includegraphics[width=\myfigwidth, height=\myfigwidth]{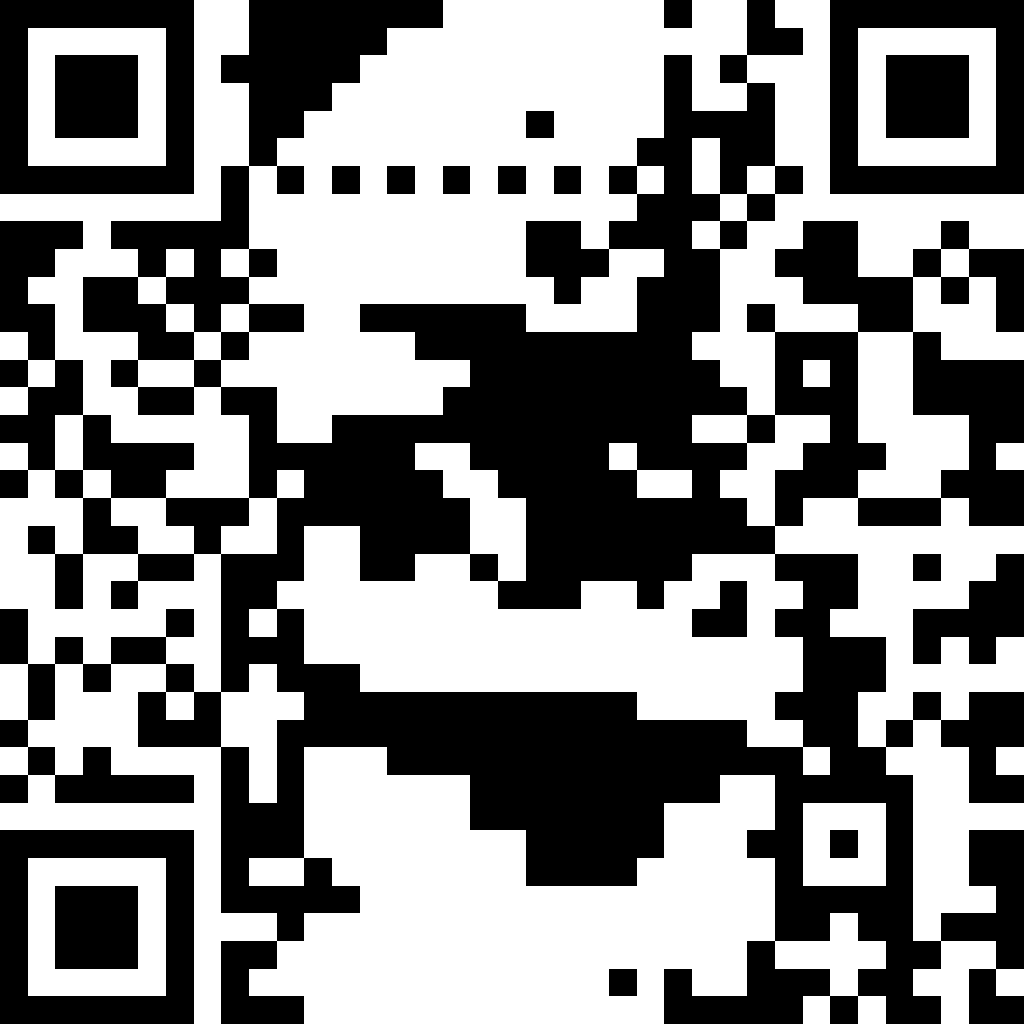} & 
    \includegraphics[width=\myfigwidth, height=\myfigwidth]{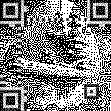} & 
    \includegraphics[width=\myfigwidth, height=\myfigwidth]{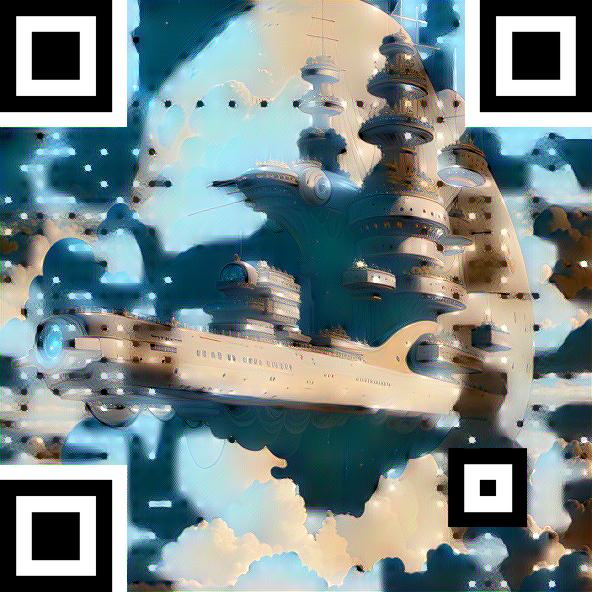} & 
    \includegraphics[width=\myfigwidth, height=\myfigwidth]{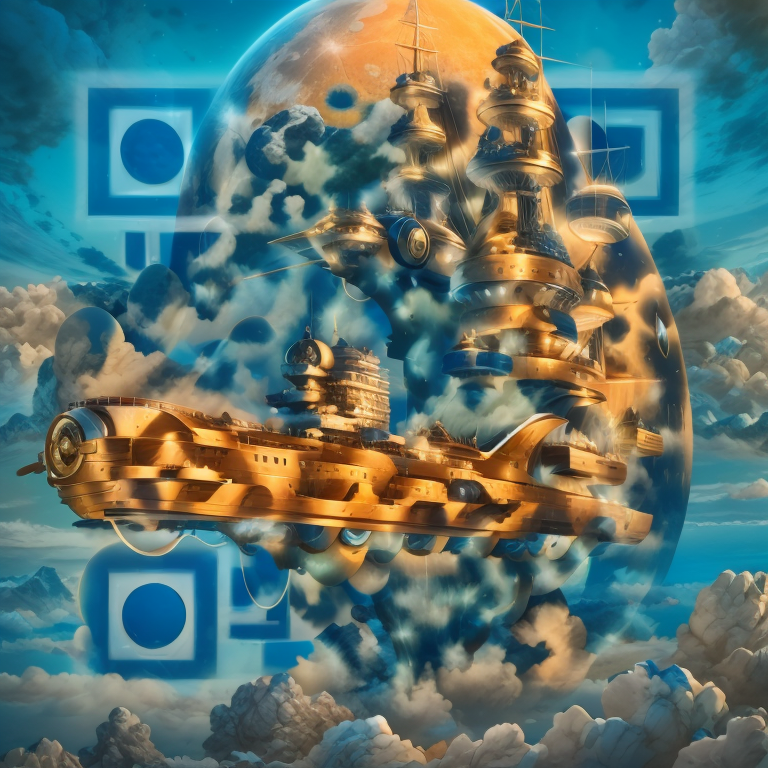} & 
    \includegraphics[width=\myfigwidth, height=\myfigwidth]{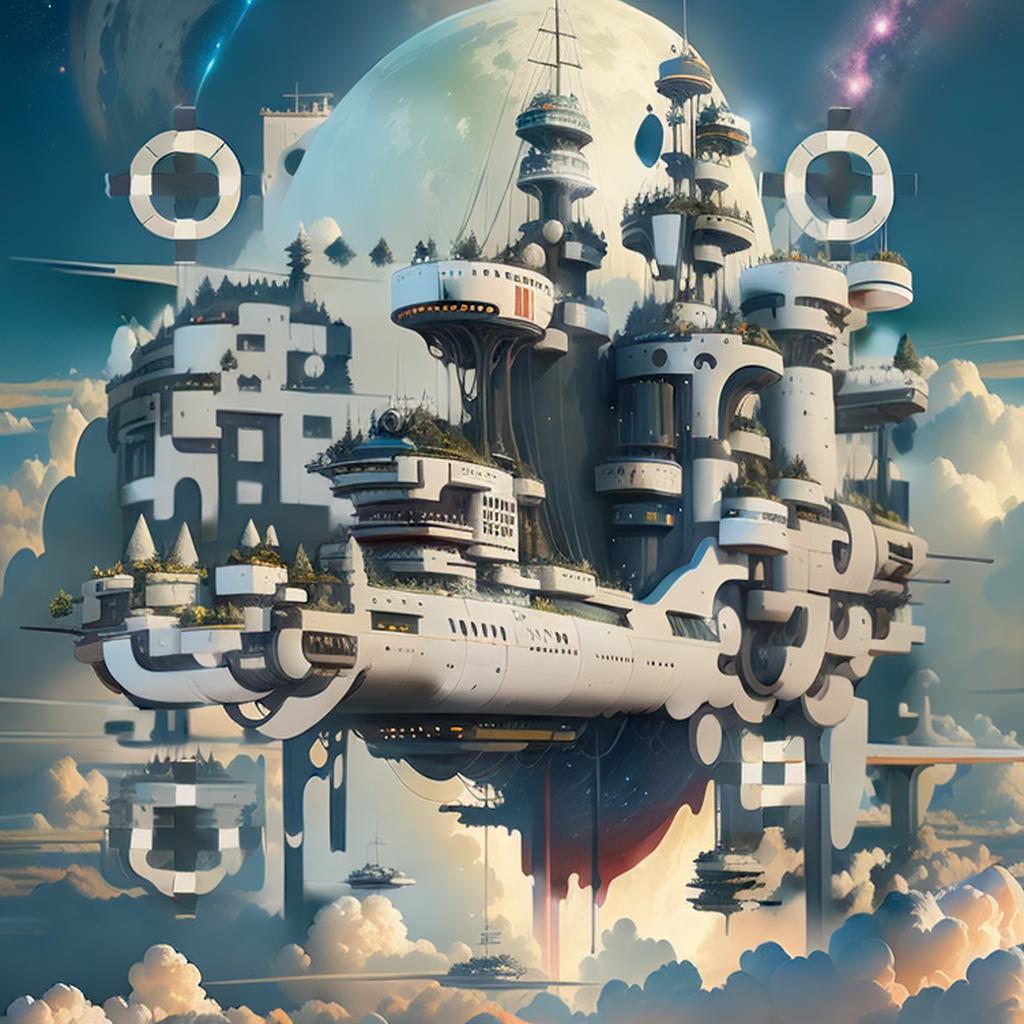} \\

    \includegraphics[width=\myfigwidth, height=\myfigwidth]{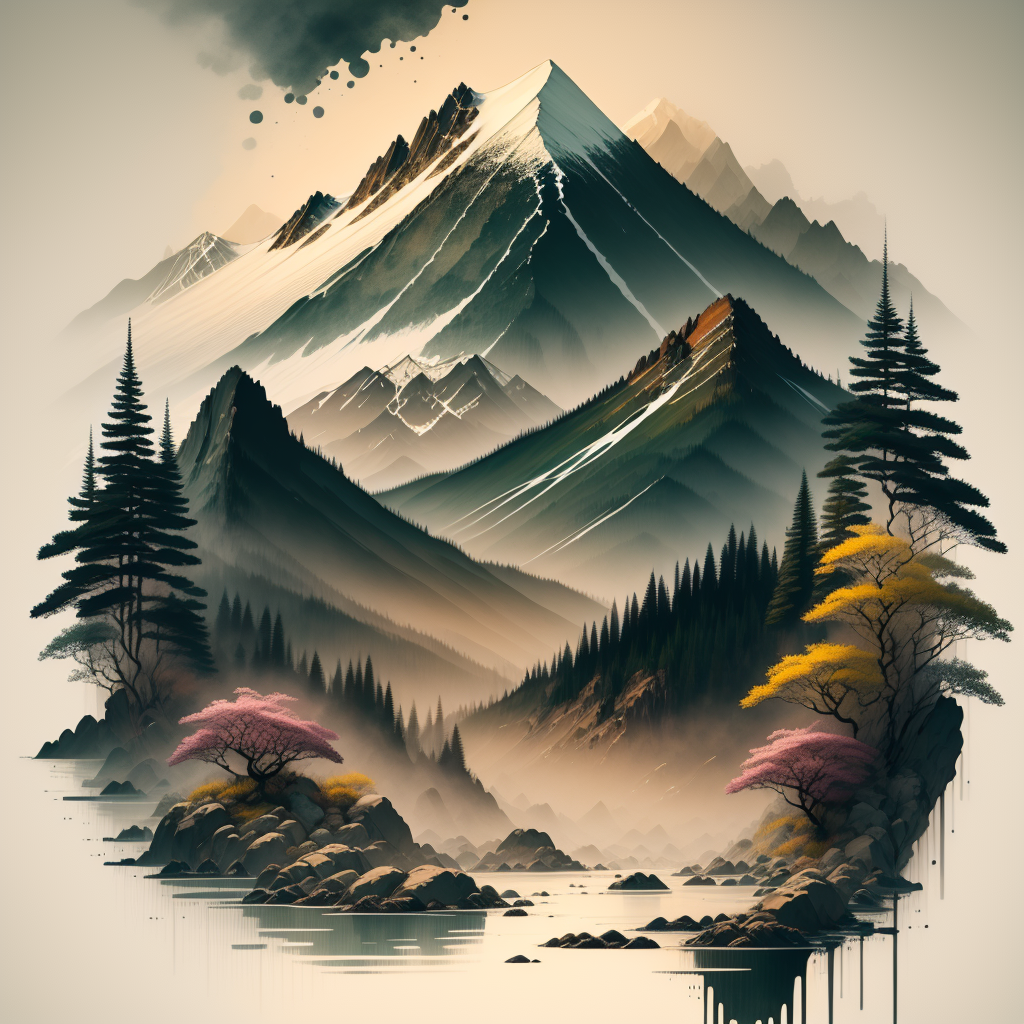} & 
    \includegraphics[width=\myfigwidth, height=\myfigwidth]{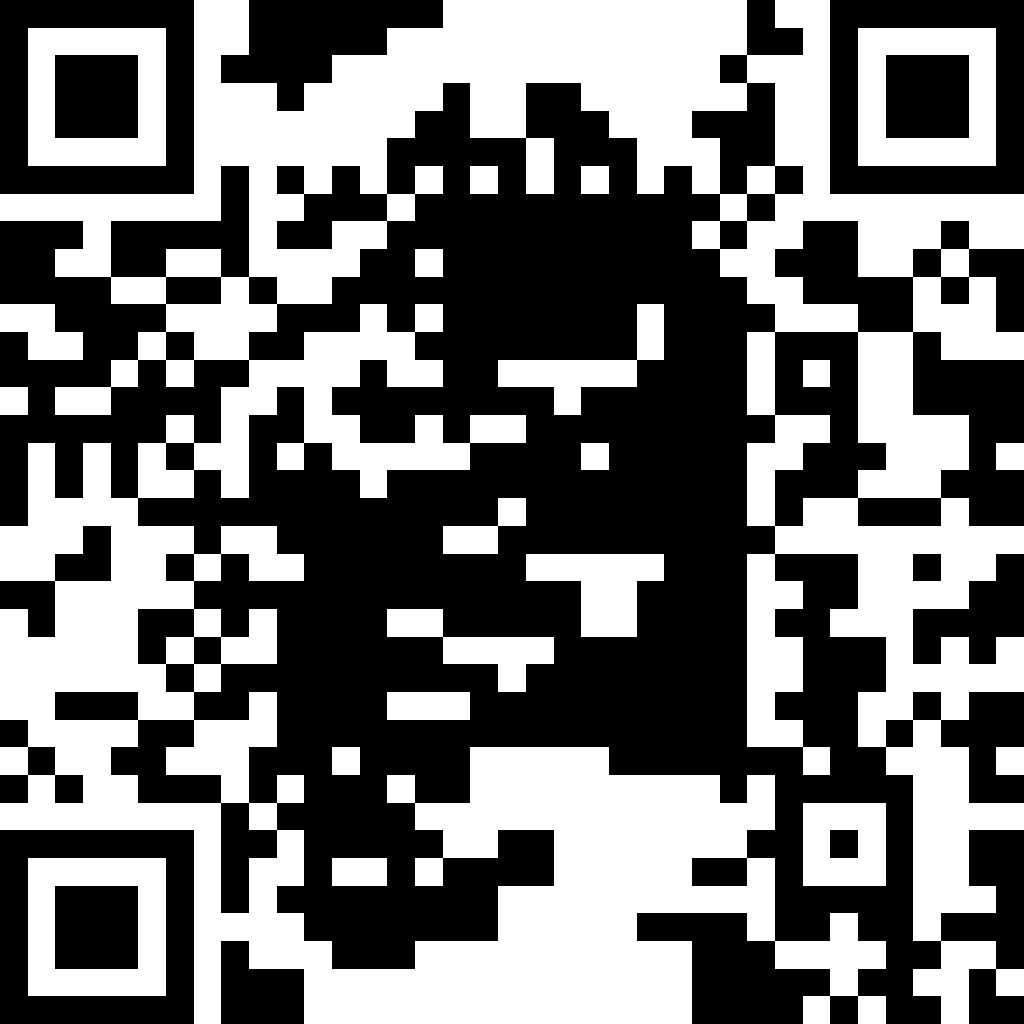} & 
    \includegraphics[width=\myfigwidth, height=\myfigwidth]{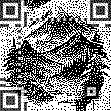} & 
    \includegraphics[width=\myfigwidth, height=\myfigwidth]{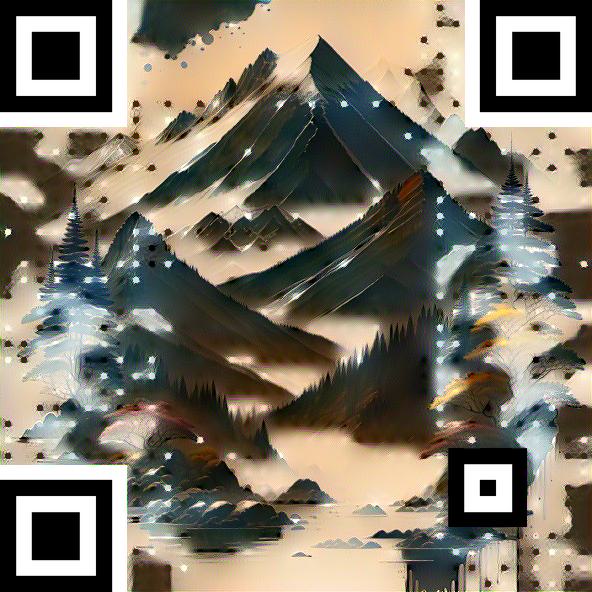} & 
    \includegraphics[width=\myfigwidth, height=\myfigwidth]{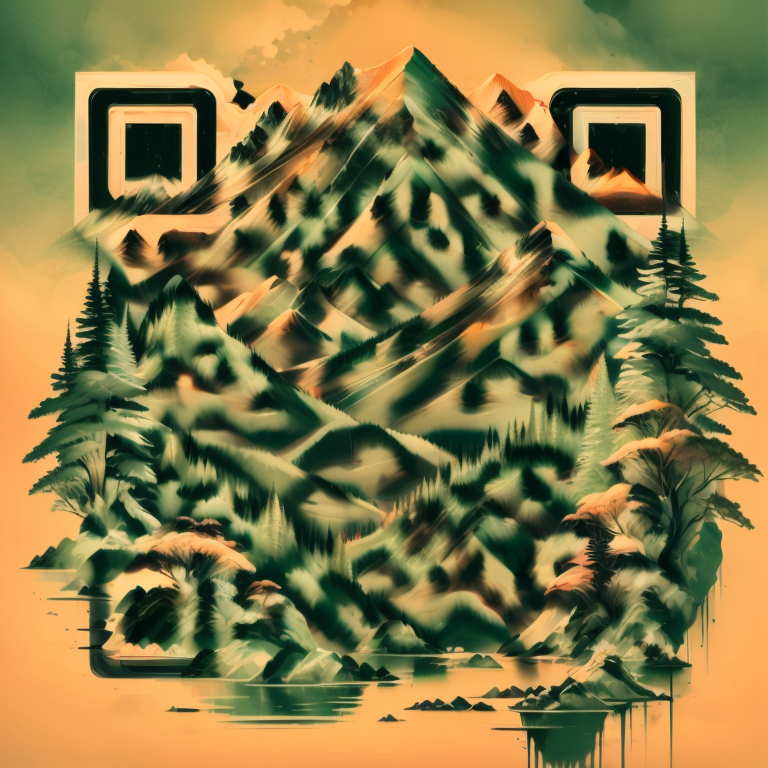} & 
    \includegraphics[width=\myfigwidth, height=\myfigwidth]{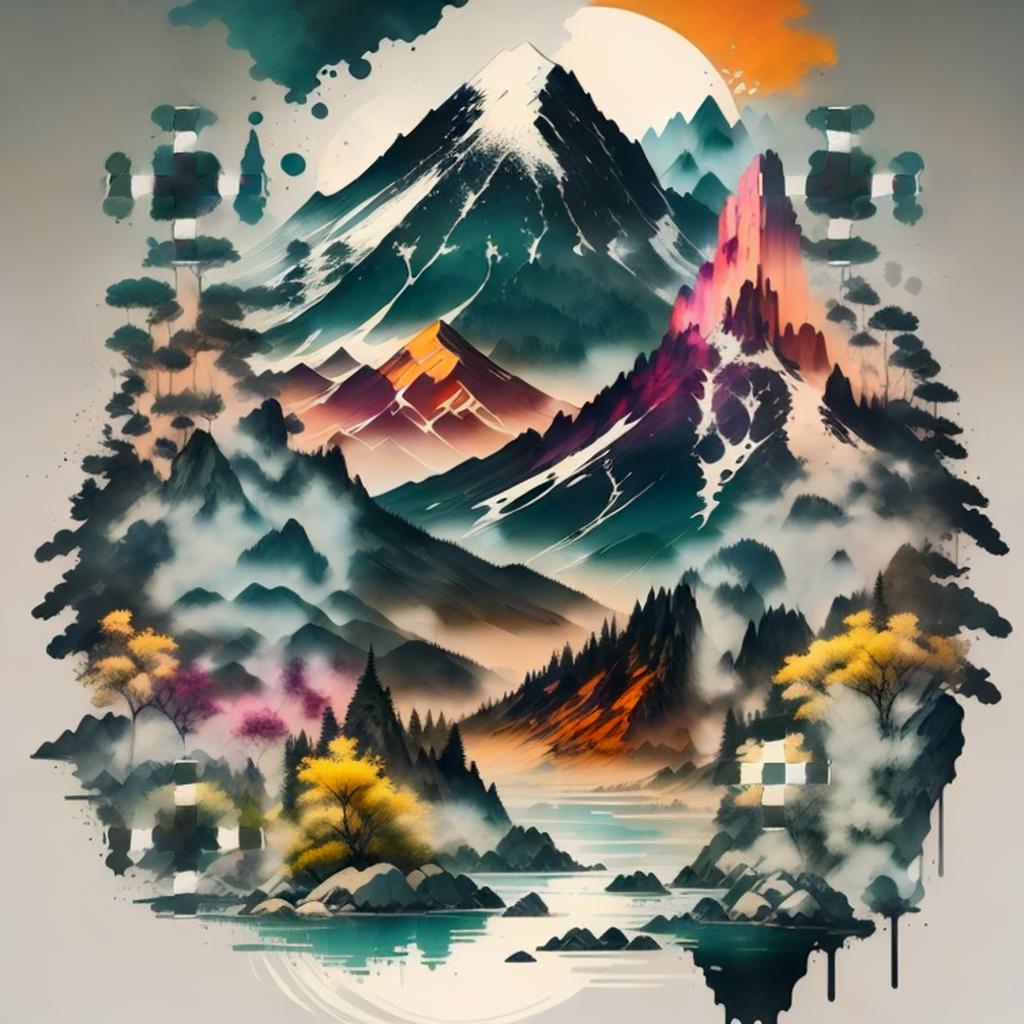} \\

    \includegraphics[width=\myfigwidth, height=\myfigwidth]{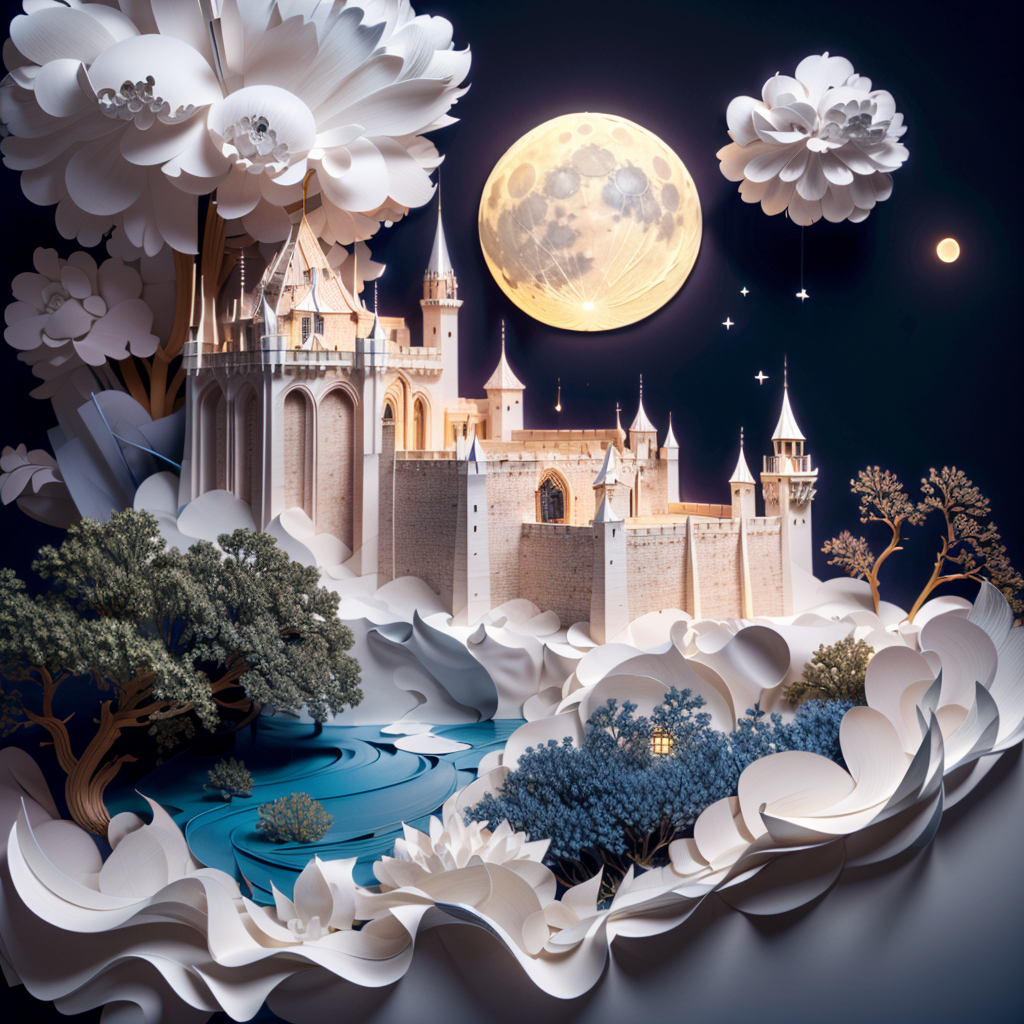} & 
    \includegraphics[width=\myfigwidth, height=\myfigwidth]{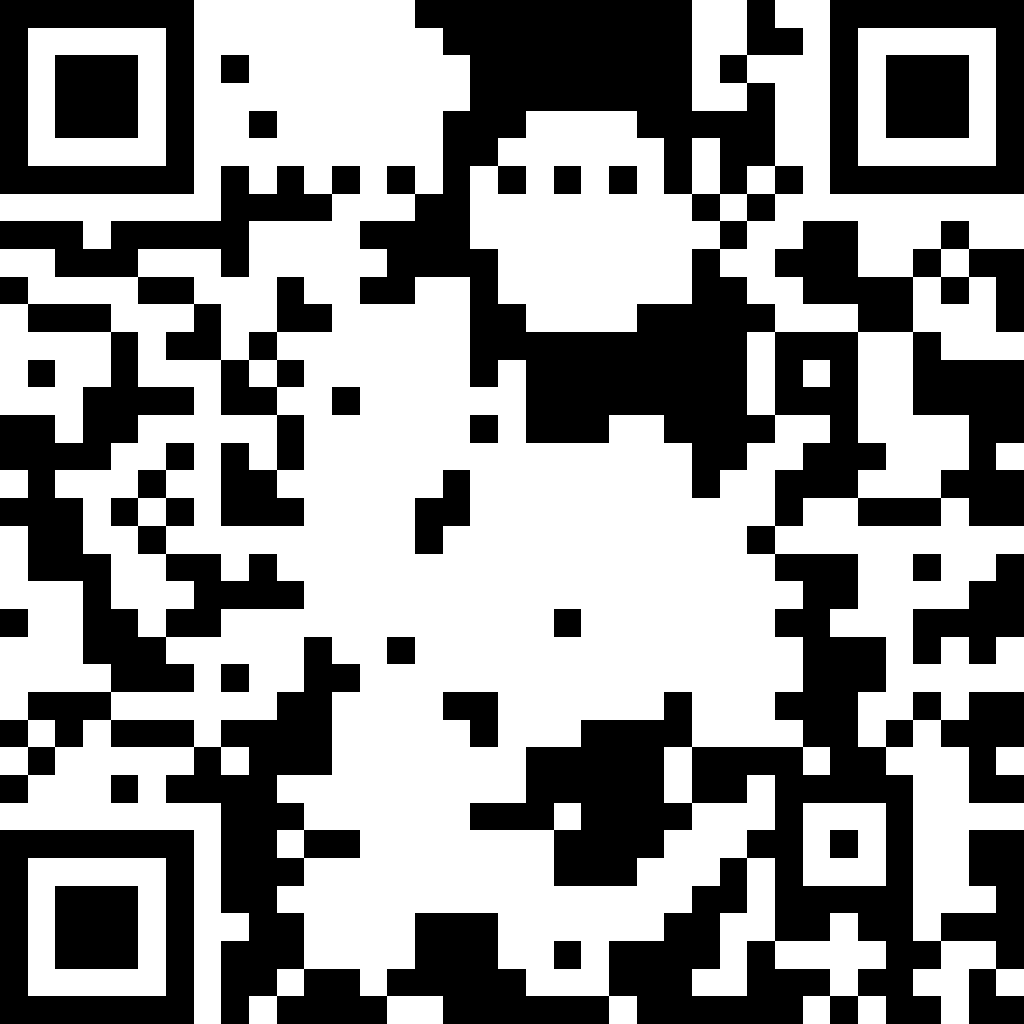} & 
    \includegraphics[width=\myfigwidth, height=\myfigwidth]{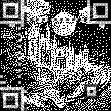} & 
    \includegraphics[width=\myfigwidth, height=\myfigwidth]{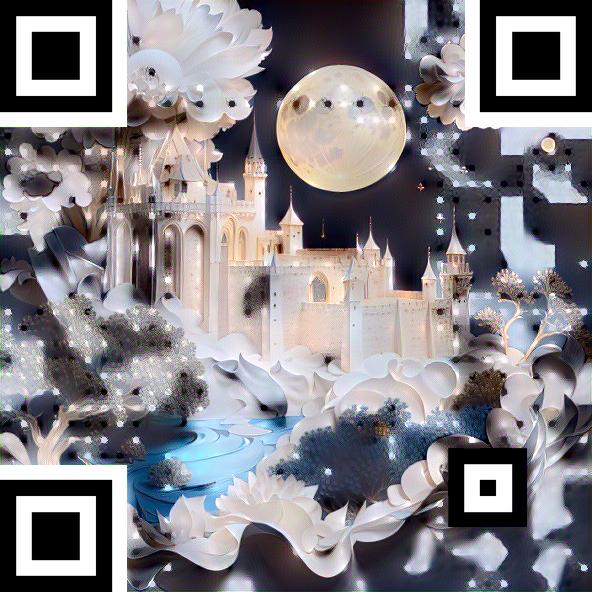} & 
    \includegraphics[width=\myfigwidth, height=\myfigwidth]{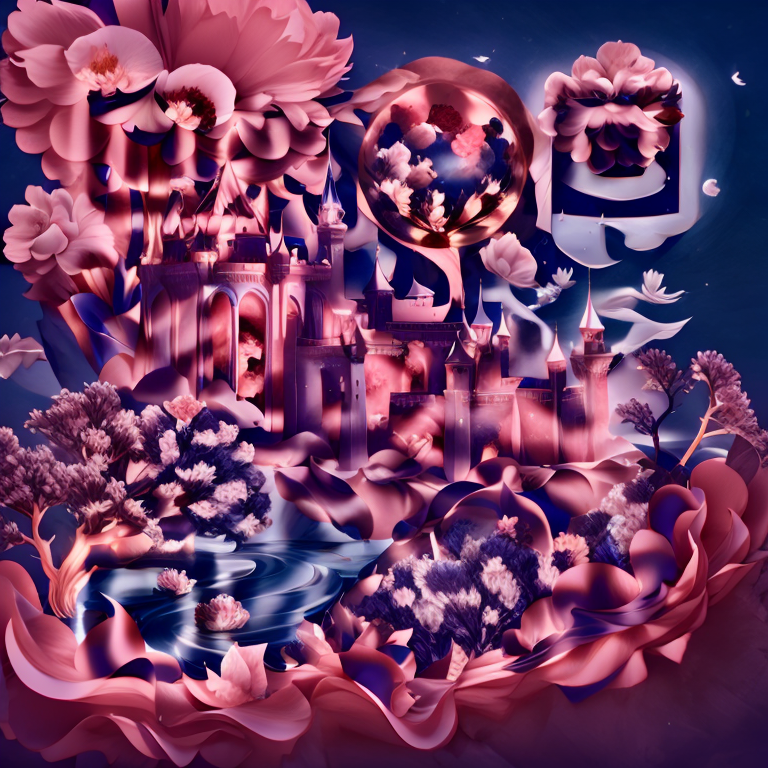} & 
    \includegraphics[width=\myfigwidth, height=\myfigwidth]{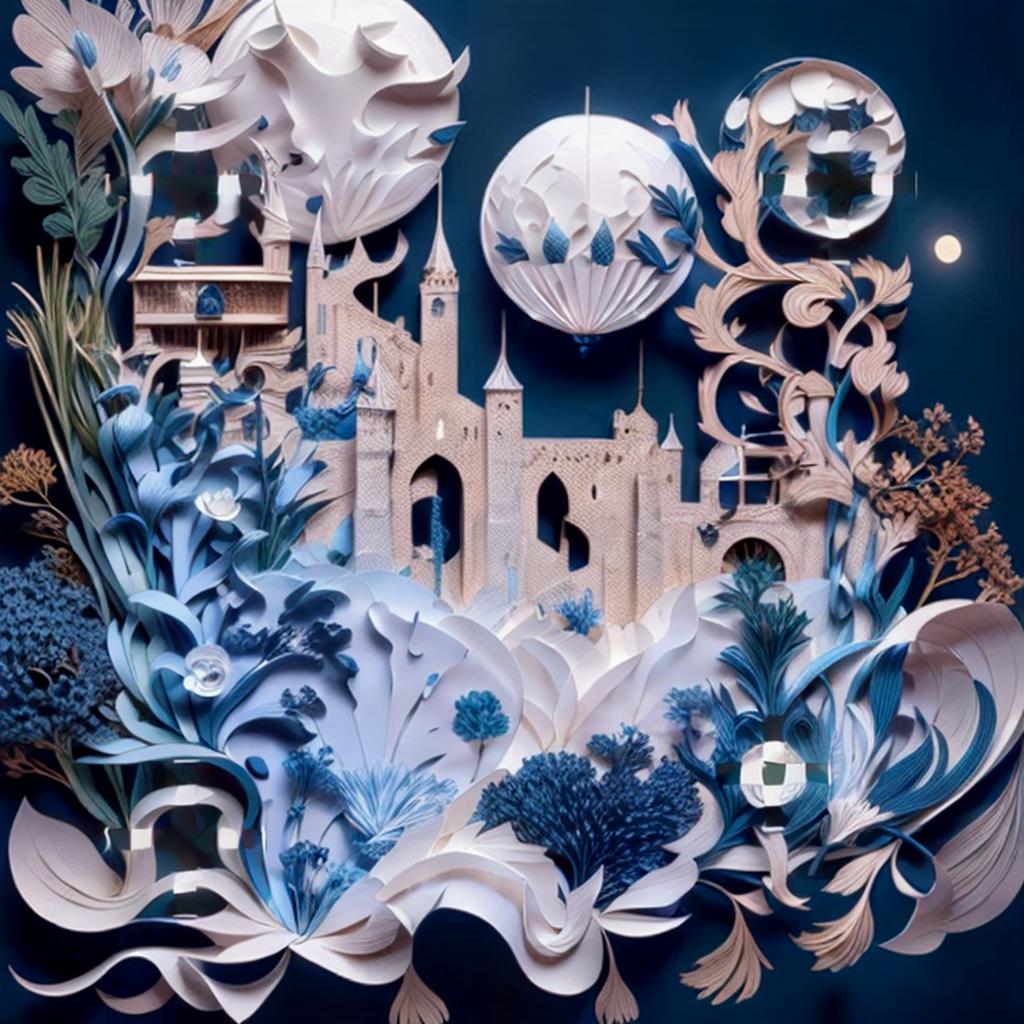} \\

    \bottomrule
    \end{tabularx}
\end{table*}

\subsection{Scanning Robustness}
We assess the performance of our QR codes across various mobile devices and readers in this paper. Initially, we generate a set of 20 aesthetic QR codes with a resolution of 512 $\times$ 512. These codes are displayed on the screen in three commonly used sizes: 3cm $\times$ 3cm, 5cm $\times$ 5cm, and 7cm $\times$ 7cm. Positioned at a distance of 20cm, we scan each code using different mobile phones and apps, varying scanning angles. We record the average number of successful scans in 50 attempts, defining success as decoding within 3 seconds.
Table~\ref{table:scannability} presents experimental results indicating that the average success rates consistently exceed 96\%. It is worth noting that even in the case where decoding exceeds 3 seconds, our QR codes are still decodeable eventually. This robust performance demonstrates the reliability of our QR codes for real-world applications.

\subsection{Aesthetic Quality}
Although scanning robustness is preserved well, we also concern the visual appeal of the QR code.

\paragraph{Comparison with existing methods.}
\begin{figure}
    \centering
    \includegraphics[width=\linewidth]{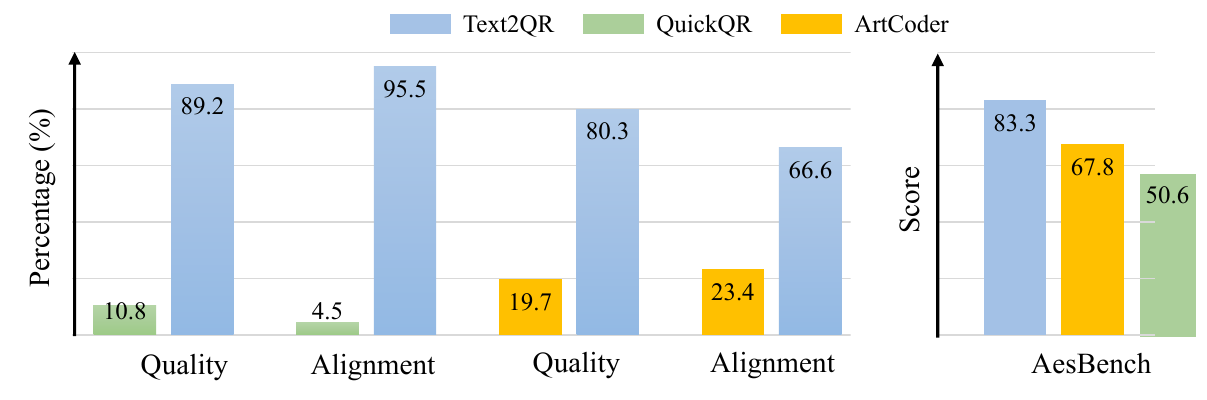}
    \caption{Statistical results of user study (left) and scores of AesBench~\cite{aesbench} (right). }
    \label{fig:user}
    \vspace{-2mm}
\end{figure}
We benchmark our methods against various aesthetic QR code approaches, including QArt~\cite{qart}, Halftone QR code~\cite{chu2013halftone}, ArtCoder~\cite{su2021artcoder}, and Quick QR~\cite{quickqr}, as detailed in Table~\ref{tab:comparison}. For ArtCoder that utilizes the neural-style transfer technique, we designate $I^g$ as both the content and style target. For QArt, Halftone QR code, and Quick QR, we employ $I^g$ as their reference input images. ArtCoder's results exhibit visible, undesired round spots, indicating repaired modules that can be distractable. Quick QR, on the other hand, yields inconsistent outcomes with the customized input. In contrast, our QR codes seamlessly integrate with the customized input image, featuring modules that are nearly invisible, ensuring superior aesthetic quality characterized by personalization, diversity, and artistic appeal.


In Figure~\ref{fig:user}, we showcase the outcomes of a user study involving 24 subjects comparing 200 generated QR-code images from various methods, where the ratio values indicate the percentages of participants preferring the corresponding model. Concurrently, we also employ AesBench~\cite{aesbench} (ranges from $0$ to $100$, the higher the better) as an Aesthetic Assessment Metric to systematically score the different methods. Our method achieves superior performance across all evaluated aspects.

\subsection{Ablation Study}
We conduct a comprehensive ablation study, validating the necessity of each module in Text2QR.

\paragraph{QAB Module.}
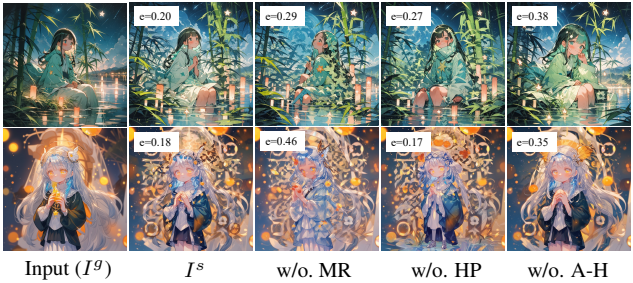
\begin{figure}[t]
    \centering
    \includegraphics[width=\linewidth]{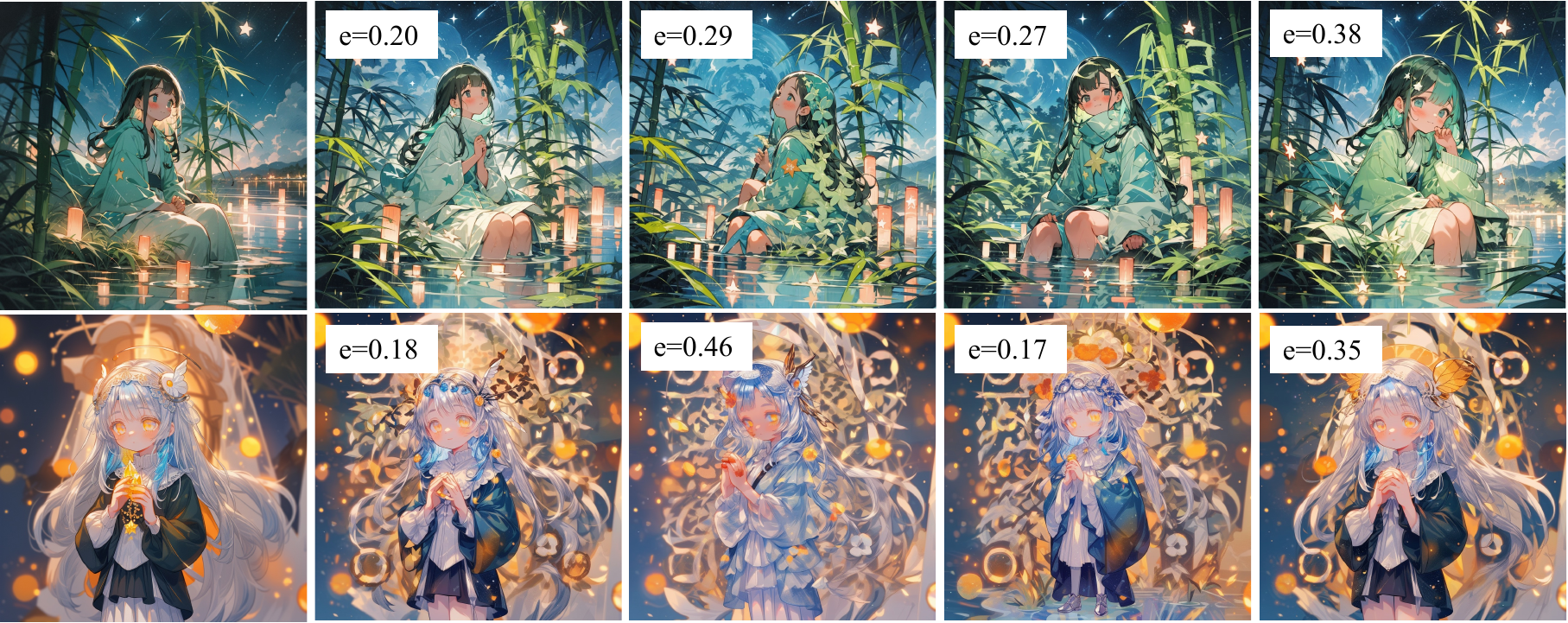}
    \begin{picture}(0,0)
        \put(-93,5){\makebox(0,0)[c]{\footnotesize Input ($I^g$)}}
        \put(-45,5){\makebox(0,0)[c]{\footnotesize $I^s$}}
        \put(0,5){\makebox(0,0)[c]{\footnotesize w/o.\@ MR}}
        \put(50,5){\makebox(0,0)[c]{\footnotesize w/o. HP}}
        \put(96,5){\makebox(0,0)[c]{\footnotesize w/o. A-H}}
    \end{picture}
    \vspace{-2mm}
    \caption{QAB Ablation Study: We assess the impact of Module Reorganization (MR), Histogram Polarization (HP), and Adaptive-Halftone Blending (A-H) on the generated $I^s$. Our result exhibits high consistency with the customized input $I^g$ and achieves a lower error level. (Note: These images are not scannable.)}\label{fig:ablation1}
    \vspace{-5pt}
\end{figure}
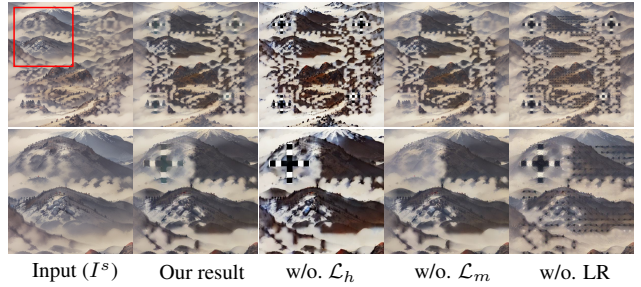
\begin{figure}[t]
    \centering
    \includegraphics[width=\linewidth]{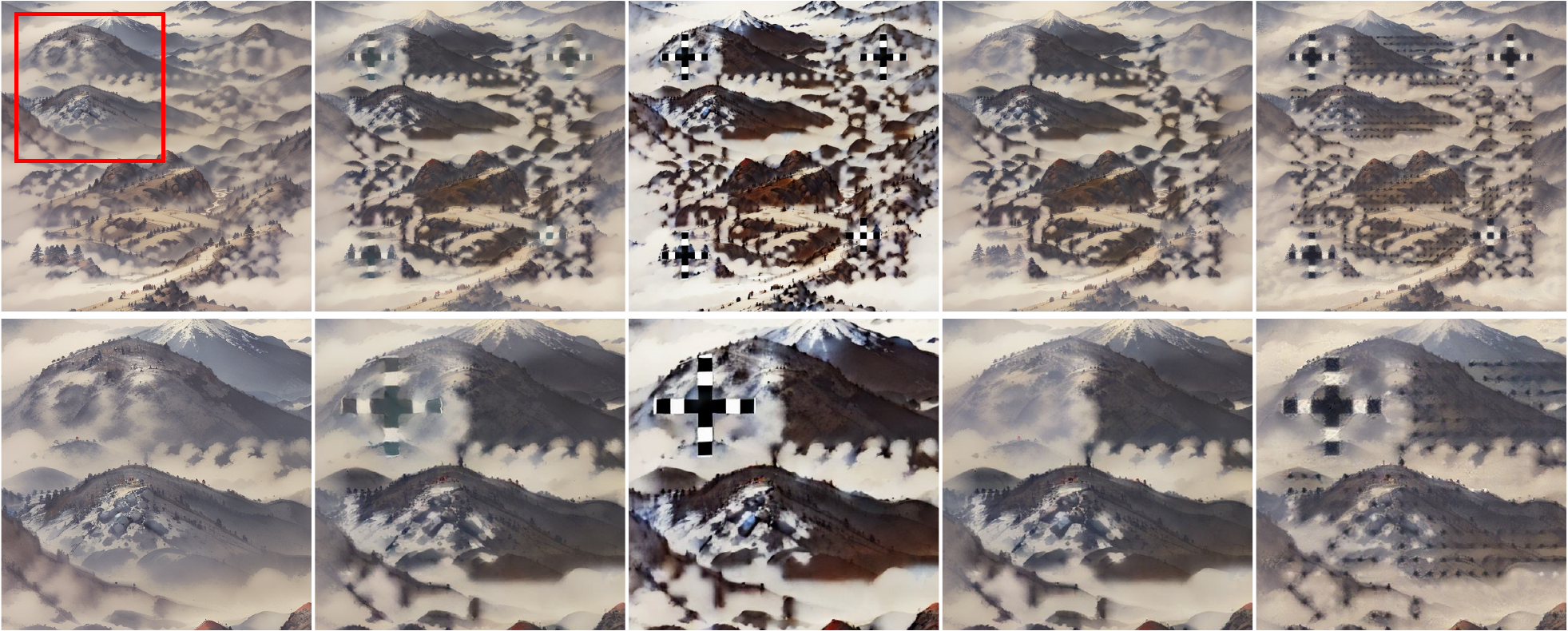}
    \begin{picture}(0,0)
        \put(-93,5){\makebox(0,0)[c]{\footnotesize Input ($I^s$)}}
        \put(-45,5){\makebox(0,0)[c]{\footnotesize Our result}}
        \put(0,5){\makebox(0,0)[c]{\footnotesize w/o.\@ $\mathcal{L}_{h}$}}
        \put(50,5){\makebox(0,0)[c]{\footnotesize w/o. $\mathcal{L}_{m}$}}
        \put(96,5){\makebox(0,0)[c]{\footnotesize w/o. LR}}
    \end{picture}
    \caption{SELR Ablation Study: The first row presents results with various losses and without Latent Refinement (LR), while the second row zooms in on the red box region.}\label{fig:compare_selr}
    \vspace{-4mm}
\end{figure}
In Text2QR, the QAB module generates a guidance blueprint image $I^b$ for the SD model to produce a high-quality aesthetic image $I^s$. Our objective is to ensure that $I^s$ not only shares a similar aesthetic style with $I^b$ but also maintains a low error level. This is achieved through module reshuffling, histogram polarization, and Adaptive-Halftone blending steps. Figure~\ref{fig:ablation1} illustrates the impact of these steps on the error level $e$ and aesthetic quality of $I^s$. Our results demonstrate a high consistency between $I^s$ and the customized input $I^g$, accompanied by a notably lower error level.

\paragraph{SELR Module.}
During the SELR process, we initialize $Q$ with $I^s$ and iteratively refine its latent code. The marker loss and code loss actively enhance scanning robustness, while the harmonizing loss meticulously controls aesthetic quality. Figure~\ref{fig:compare_selr} illustrates the importance of SELR module. Results refined directly on $Q$ (denoted as ``w/o.\@ LR'') display visible, undesired round spots. Outputs without the marker loss resemble standard outputs but are unscannable. Omitting the harmonizing loss results in outputs with discordant appearances, underscoring its crucial role in achieving harmonious results. In conclusion, SELR refines the QR code result, ensuring a harmonious blend of functionality and aesthetic quality.

\section{Conclusion}\label{sec:conclusion}

In summary, Text2QR utilizes the Stable-Diffusion (SD) model to effectively address the dual challenge of achieving user-defined aesthetics and scanning robustness in QR code generation. The strategic integration of the QR Aesthetic Blueprint (QAB) module that ensures generation stability and the Scannability Enhancing Latent Refinement (SELR) process that iteratively operates in the latent space, enhancing scanning robustness of the output. This innovative approach adeptly balances image aesthetics and scanning robustness, showcasing visual appeal and practical utility and surpassing previous approaches by a large margin, marking a substantial advancement in QR code generation.

\section{Acknowledgment}
The work was supported in part by the National Natural Science Foundation of China under Grant 62301310, and in part by the Shanghai Pujiang Program under Grant 22PJ1406800.

\clearpage\newpage
{
    \small
    \bibliographystyle{ieeenat_fullname}
    \bibliography{main}
}



\end{document}